\documentclass[nohyperref]{article}

\usepackage{microtype}
\usepackage{graphicx}
\usepackage{booktabs} 
\usepackage{courier}

\usepackage{hyperref}

\usepackage[accepted]{icml2023}

\usepackage{amsmath}
\usepackage{amssymb}
\usepackage{mathtools}
\usepackage{amsthm}

\usepackage{hyperref}
\usepackage{url}
\usepackage{booktabs}
\usepackage{graphicx}
\usepackage{svg}
\usepackage{algorithm} 
\usepackage{algorithmic} 
\usepackage{dsfont}
\usepackage{wrapfig}
\usepackage{subcaption}

\usepackage{amsmath,amsfonts,bm}

\def\eqref#1{equation~\ref{#1}}

\def\1{\bm{1}}

\def\eps{{\epsilon}}

\def\rmI{{\mathbf{I}}}

\def\vmu{{\bm{\mu}}}

\def\veps{{\bm{\eps}}}

\def\vv{{\bm{v}}}

\def\vx{{\bm{x}}}

\def\vz{{\bm{z}}}

\DeclareMathAlphabet{\mathsfit}{\encodingdefault}{\sfdefault}{m}{sl}
\SetMathAlphabet{\mathsfit}{bold}{\encodingdefault}{\sfdefault}{bx}{n}

\usepackage{listings,newtxtt}

\definecolor{deepblue}{rgb}{0,0,0.6}
\definecolor{deepred}{rgb}{0.6,0,0}
\definecolor{deepgreen}{rgb}{0,0.5,0}
\lstdefinestyle{python}{
language=Python,
basicstyle=\ttfamily\small,
commentstyle=\color{deepred},
otherkeywords={},             
keywordstyle=\color{deepgreen},
emph={},          
emphstyle=\color{deepblue},    
stringstyle=\color{deepred},
showstringspaces=false            %
}

\usepackage{courier}

\usepackage[capitalize,noabbrev]{cleveref}

\theoremstyle{plain}

\theoremstyle{definition}

\theoremstyle{remark}

\usepackage[textsize=tiny]{todonotes}

\icmltitlerunning{simple diffusion}

\begin{document}

\twocolumn[
\icmltitle{simple diffusion: End-to-end diffusion for high resolution images}



\icmlsetsymbol{equal}{*}

\begin{icmlauthorlist}
\icmlauthor{Emiel Hoogeboom}{equal,goog}
\icmlauthor{Jonathan Heek}{equal,goog}
\icmlauthor{Tim Salimans}{goog}
\end{icmlauthorlist}

\icmlaffiliation{goog}{Google Research, Brain Team, Amsterdam, Netherlands}

\icmlcorrespondingauthor{Emiel Hoogeboom}{emielh@google.com}

\icmlkeywords{Machine Learning, ICML}

\vskip 0.3in
]



\printAffiliationsAndNotice{\icmlEqualContribution} 

\begin{abstract}
Currently, applying diffusion models in pixel space of high resolution images is difficult. Instead, existing approaches focus on diffusion in lower dimensional spaces (latent diffusion), or have multiple super-resolution levels of generation referred to as cascades. The downside is that these approaches add additional complexity to the diffusion framework.

This paper aims to improve denoising diffusion for high resolution images while keeping the model as simple as possible. The paper is centered around the research question: How can one train standard diffusion models on high resolution images, and still obtain performance comparable to these alternate approaches? 

The four main findings are: 1) the noise schedule should be adjusted for high resolution images, 2) It is sufficient to scale only a particular part of the architecture, 3) dropout should be added at specific locations in the architecture, and 4) downsampling is an effective strategy to avoid high resolution feature maps. Combining these simple yet effective techniques, we achieve state-of-the-art on image generation among diffusion models without sampling modifiers on ImageNet.
\end{abstract}

\begin{figure}
    \centering
    \includegraphics[width=.21\textwidth]{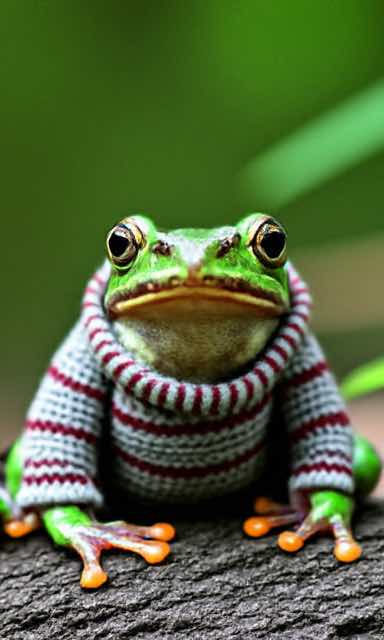} \hspace{.01\textwidth}
    \includegraphics[width=.21\textwidth]{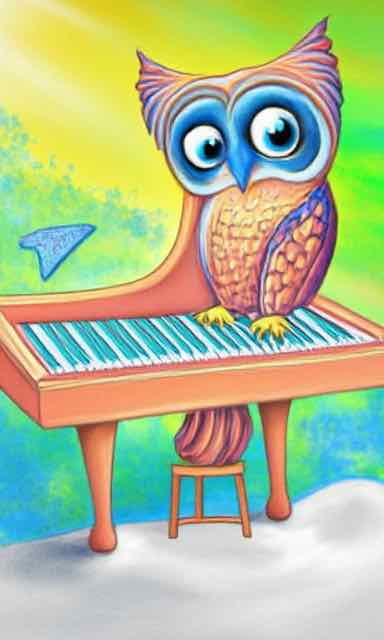} \\
    \vspace{.24cm}
    \includegraphics[width=.44\textwidth]{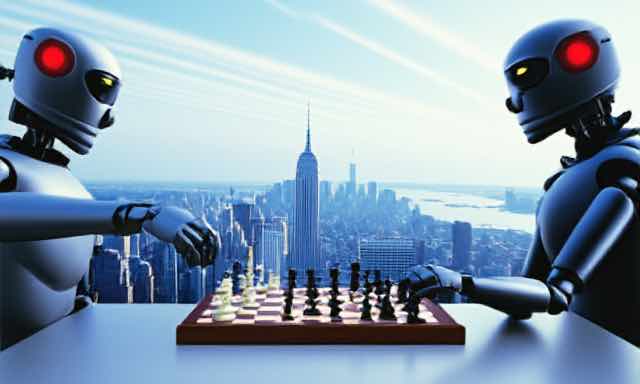}
    \caption{\textit{A dslr photo of a frog wearing a sweater}, \textit{An owl playing the piano, vivid, fantasy art}, and \textit{two robots playing chess with New York in the background}. Except for the frozen text encoder, \textit{simple diffusion} is trained end-to-end and images are generated in full pixel space.}
    \label{fig:frog_sweater}
    \vspace{-.4cm}
\end{figure}

\section{Introduction}
Score-based diffusion models have become increasingly popular for data generation. In essence the idea is simple: one pre-defines a diffusion process, which gradually destroys information by adding random noise. Then, the opposite direction defines the denoising process, which is approximated with a neural network.\vspace{.2cm}

Diffusion models have shown to be extremely effective for image, audio, and video generation. However, for higher resolutions the literature typically operates on lower dimensional latent spaces (latent diffusion) \citep{rombach2022highresolution} or divides the generative process into multiple sub-problems, for instance via super-resolution (cascaded diffusion) \citep{ho2022cascaded} or mixtures-of-denoising-experts \citep{balaji2022ediffi}. The disadvantage is that these approaches introduce additional complexity and usually do not support a single end-to-end training setup.

\begin{figure*}[]
    \centering
    \includegraphics[width=.9\textwidth]{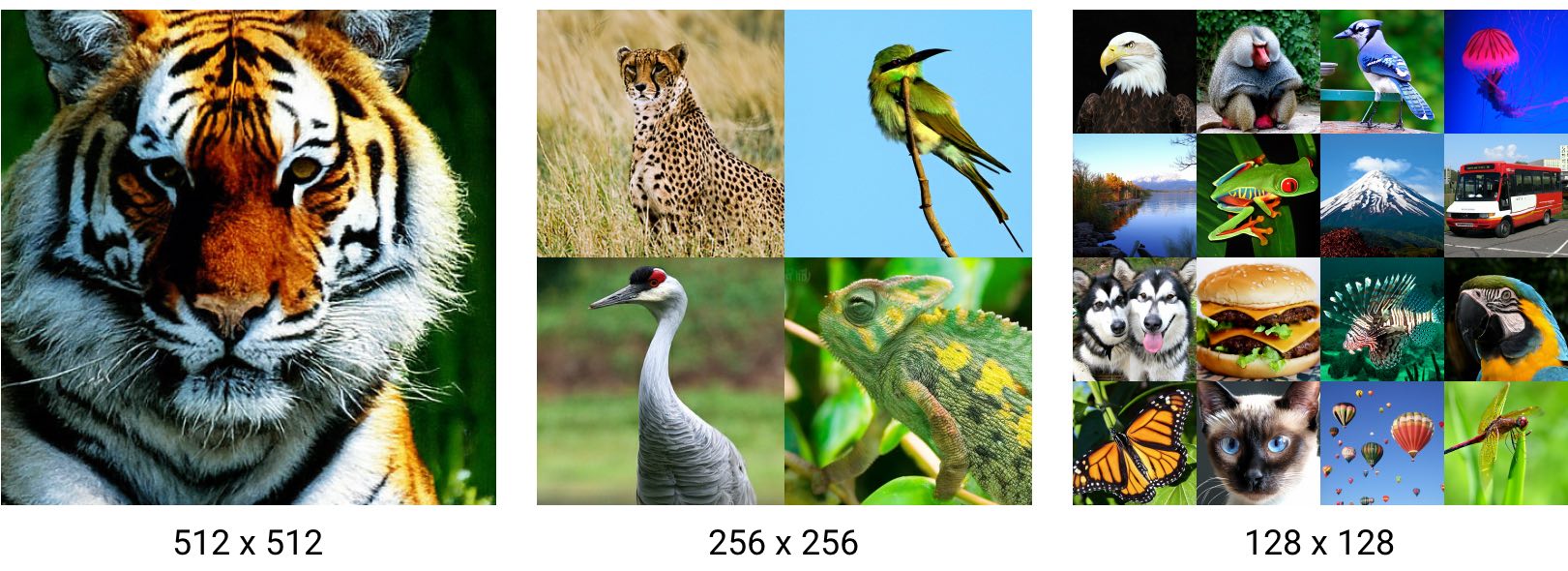}\vspace{-.2cm}
    \caption{Generated images with \textit{simple diffusion}. Importantly, each image is generated in full image space by a single diffusion model without any cascades (super-resolution) or mixtures of experts. Samples are drawn from the U-Net model with guidance scale 4.}
    \label{fig:overview}
\end{figure*}

In this paper, we aim to improve standard denoising diffusion for higher resolutions while keeping the model as simple as possible. Our four main findings are that 1) the noise schedule should be adjusted for larger images,  adding more noise as the resolution increases. 2) It is sufficient to scale the U-Net architecture on the $16 \times 16$ resolution to improve performance. Taking this one step further is the U-ViT architecture, a U-Net with a transformer backbone. 3) Dropout should be added for improved performance, but not on the highest resolution feature maps. And finally 4) for higher resolutions, one can down-sample without performance degradation. 
Most importantly, these results are obtained using just a single model and an end-to-end training setup. After using existing distillation techniques which now only have to be applied to a single stage, the model can generate an image in 0.4 seconds.

\section{Background: Diffusion Models}
\label{sec:background}

A diffusion model generates data by learning the reverse of a destruction process. Commonly, the diffusion process gradually adds Gaussian noise over time. It is convenient to express the process directly in the marginals $q(\vz_t | \vx)$ which is given by:
\begin{equation}\small
    q(\vz_t | \vx) = \mathcal{N}(\vz_t | \alpha_t \vx, \sigma_t^2 \rmI)
\end{equation}
where $\alpha_t, \sigma_t \in (0, 1)$ are hyperparameters that determine how much signal is destroyed at a timestep $t$, which can be continuous for instance $t \in [0, 1]$. Here, $\alpha_t$ is decreasing and $\sigma_t$ is increasing, both larger than zero. We consider a variance preserving process, which fixes the relation between $\alpha_t, \sigma_t$ to be $\alpha_t^2 = 1 - \sigma_t^2$. Assuming the diffusion process is Markov, the transition distributions are given by:
\begin{equation}\small
    q(\vz_t | \vz_s) = \mathcal{N}(\vz_t | \alpha_{ts} \vz_s, \sigma_{ts}^2 \rmI)
\end{equation}
where $\alpha_{ts} = \alpha_t / \alpha_s$ and $\sigma_{ts}^2 = \sigma_t^2 - \alpha_{t|s}^2 \sigma_s^2$ and $t > s$.

\textbf{Noise schedule}
An often used noise schedule is the $\alpha$-cosine schedule where $\alpha_t = \cos(\pi t / 2)$ which under the variance preserving assumption implies $\sigma_t = \sin(\pi t / 2)$. An important finding from \citep{kingma2021vdm} is that it is the signal-to-noise ratio $\alpha_t / \sigma_t$ that matters, which is then $1 / \tan(\pi t / 2)$ or in log space $\log \frac{\alpha_t}{\sigma_t} = -\log \tan(\pi t / 2)$.

\textbf{Denoising}
Conditioned on a single datapoint $\vx$, the denoising process can be written as:
\begin{equation}\small
    q(\vz_s | \vz_t, \vx) = \mathcal{N}(\vz_t | \vmu_{t \to s}, \sigma_{t \to s}^2 \rmI).
\end{equation}
where $\vmu_{t \to s} = \frac{\alpha_{ts} \sigma_s^2}{\sigma_t^2} \vz_t + \frac{\alpha_s \sigma_{ts}^2}{\sigma_t^2} \vx$ and $ \sigma_{t \to s} = \frac{\sigma_{ts}^2 \sigma_{s}^2}{\sigma_t^2}$. An important and surprising result in literature is that when $\vx$ is approximated by a neural network $\hat{\vx} = f_\theta(\vz_t)$, then one can define the learned distribution $p(\vz_s | \vz_t) = q(\vz_s | \vz_t, \vx = \hat{\vx})$ without loss of generality as $s \to t$. This works because as $s \to t$, the true denoising distribution for all datapoints $q(\vz_s | \vz_t)$ (which is typically unknown) will become equal to $q(\vz_s | \vz_t, \vx = \mathbb{E}[\vx | \vz_t])$ \citep{song2021scorebasedsde}.

\textbf{Parametrization}
The network does not need to approximate $\hat{\vx}$ directly, and experimentally it has been found that other predictions produce higher visual quality. Studying the re-parametrization of the marginal $q(\vz_t | \vx)$ which is $\vz_t = \alpha_t \vx + \sigma_t \veps_t$ where $\veps_t \sim \mathcal{N}(0, 1)$, one can for instance choose the \textit{epsilon} parametrization where the neural net predicts $\hat{\veps}_t$. To obtain $\hat{\vx}$, one computes $\hat{\vx} = \vz_t / \alpha_t - \sigma_t \hat{\veps}_t / \alpha_t$. The problem with the epsilon parametrization is that it gives unstable sampling near $t = 1$. An alternative parametrization without this issue is called \textit{v prediction} and was proposed in \citep{salimans2022progressive}, it is defined as $\hat{\vv}_t = \alpha_t \hat{\veps}_t - \sigma_t \hat{\vx}$.

Note that given $\vz_t$ one can obtain $\hat{\vx}$ and $\hat{\veps}_t$ via the identities $\sigma_t \vz_t + \alpha_t \hat{\vv}_t = (\sigma_t^2 + \alpha_t^2) \hat{\veps}_t = \hat{\veps}_t $ and  $\alpha_t \vz_t - \sigma_t \hat{\vv}_t = (\alpha_t^2 + \sigma_t^2) \hat{\vx} = \hat{\vx}$. In initial experiments we found \textit{v prediction} to train more reliably, especially for larger resolutions, and therefore we use this parametrization throughout this paper.

\begin{figure*}
    \centering
    \includegraphics[interpolate=false,width=.7\textwidth]{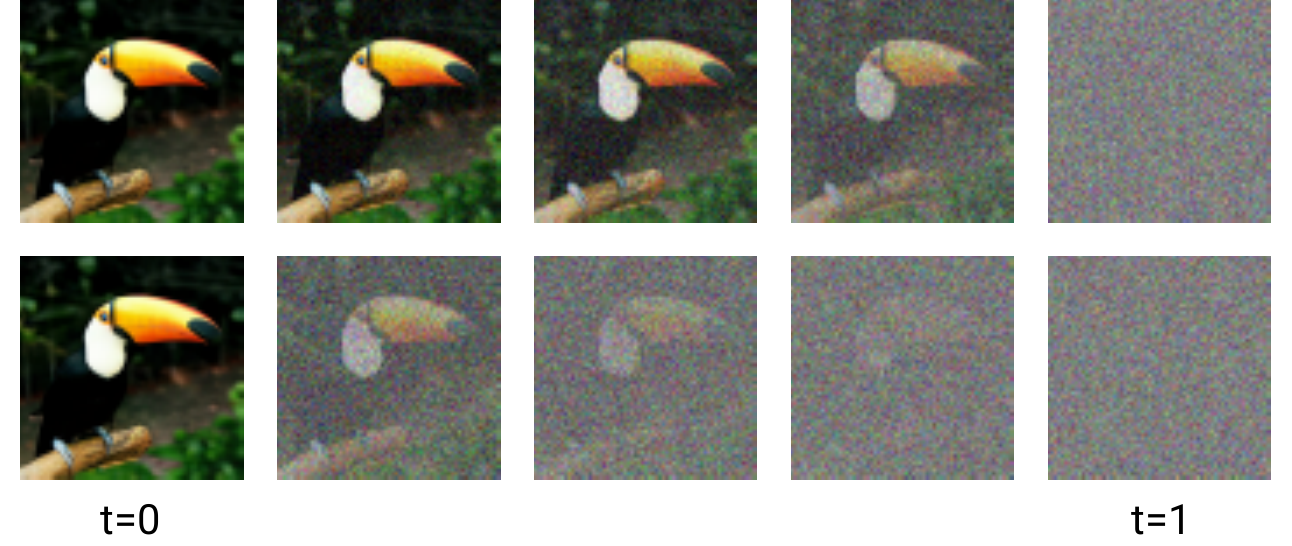}
    \includegraphics[width=.235\textwidth]{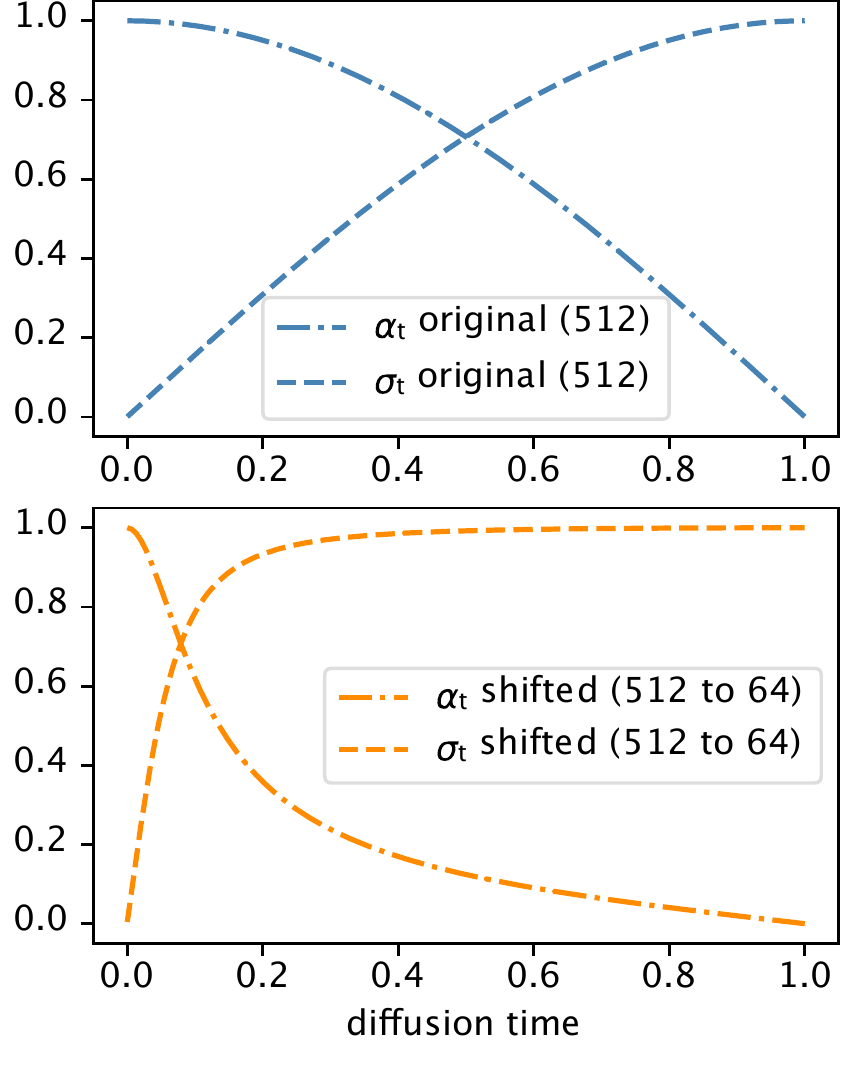}
    \vspace{-.2cm}
    \caption{The standard and shifted diffusion noise on an image of $512 \times 512$, that is visualized by average pooling to a resolution of $64 \times 64$. The top row shows a conventional cosine schedule, the bottom row shows our proposed shifted schedule.}
    \label{fig:diffusion_shifted}
\end{figure*}

\textbf{Optimization}
To train the model, we use the standard epsilon loss from \citep{ho2020denoising}. A way to motivate this choice of loss, is that using variational inference one can derive a lowerbound (in continuous time) on the model log-likelihood as done in \citep{kingma2021vdm}:
\small\begin{align*}
\begin{split}
   \log p(\vx) &= \log \mathbb{E}_{q} \frac{p(\vx, \vz_0, \ldots, \vz_1)}{q(\vz_0, \ldots, \vz_1 | \vx)} \geq \mathbb{E}_{q} \log \frac{p(\vx, \vz_0, \ldots, \vz_1)}{q(\vz_0, \ldots, \vz_1 | \vx)} \\
   &= \mathcal{L}_x + \mathcal{L}_T - \mathbb{E}_{t \sim \mathcal{U}(0, 1)} \Big{[} w(t) ||\veps_t - \hat{\veps}_t ||^2 \Big{]}, 
\end{split}
\end{align*}\normalsize
where for a well-defined process $\mathcal{L}_x = -\log p(\vx | \vz_0) \approx 0$ for discrete $\vx$, $\mathcal{L}_T = -\mathrm{KL}(q(\vz_T | \vx) | p(\vz_T)) \approx 0$, and where $w(t)$ is a weighting function which for the equation to be true needs to be $w(t) = - \frac{\mathrm{d}}{\mathrm{d}t}\log \mathrm{SNR}(t)$ where $\mathrm{SNR}(t) = \alpha_t^2 / \sigma_t^2$. In practice, we generally use the unweighted loss on $\veps_t$ (meaning that $w(t) = 1$) which in \citep{ho2020denoising} was found to give superior sample quality. See Appendix~\ref{sec:addition_info_diffusion} for additional useful background information.

\section{Method: simple diffusion}
In this section, we introduce several modifications that enable denoising diffusion to work well on high resolutions.

\subsection{Adjusting Noise Schedules}
One of the modifications is the noise schedule that is typically used for diffusion models. The most common schedules is the $\alpha$-cosine schedule, which under the variance preserving assumption amounts to $\frac{\sigma_t}{\alpha_t} = \tan(\pi t / 2)$ (ignoring the boundaries around $t=0$ and $t = 1$ for this analysis) \citep{nichol2021improvedddpm}. This schedule was originally proposed to improve the performance on CIFAR10 which has a resolution of $32 \times 32$ and ImageNet $64 \times 64$.

However, for high resolutions not enough noise is added. For instance, inspecting the top row of Figure~\ref{fig:diffusion_shifted} shows that for the standard cosine schedule, the global structure of the image is largely defined already for a wide range in time. This is problematic because the generative denoising process only has a small time window to decide on the global structure of the image. We argue that for higher resolutions, this schedule can be changed in a predictable way to retain good visual sample quality.

To illustrate this need in more detail, let us study a $128 \times 128$ problem. Given an input image $\vx$ the diffusion distribution for pixel $i$ is given by $q(z_t^{(i)} | \vx) = \mathcal{N}(z_t^{(i)} | \alpha_t x_i, \sigma_t)$. Commonly, diffusion models use network architectures that use \textit{downsampling} to operate on lower resolution feature maps, in our case with average pooling. Suppose we average pool $\vz_t$, where we let indices $1, 2, 3, 4$ denote the pixels in a $2 \times 2$ square that is being pooled. This new pixel is $z^{64 \times 64}_t = (z_t^{(1)} + z_t^{(2)} + z_t^{(3)} + z_t^{(4)}) / 4$. Recall that for variance of independent random variables is additive meaning that $\mathrm{Var}[X_1 + X_2] = \mathrm{Var}[X_1] + \mathrm{Var}[X_2]$ and that $\mathrm{Var}[aX] = a^2\mathrm{Var}[X]$ for a constant $a$. Letting $x^{64 \times 64}$ denote the first pixel of the average pooled input image, we find that $z^{64 \times 64}_t \sim \mathcal{N}(\alpha_t x^{64 \times 64}, \sigma_t / 2)$. The lower resolution pixel $z^{64 \times 64}_t$ only has half the amount of noise. We hypothesize that as resolutions increase this is problematic, as much less diffusion time is spent on the lower resolution, a stage at which the global consistency is generated.

\begin{figure*}
\centering
\begin{subfigure}[t]{0.33\textwidth}
\centering
\includegraphics[width=.99\textwidth]{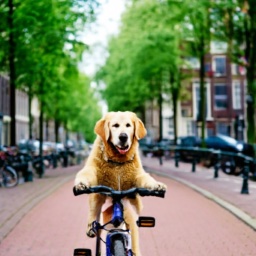}
\caption{\textit{A dog riding a bicycle through Amsterdam}}
\end{subfigure} \hfill
\begin{subfigure}[t]{0.33\textwidth}
\centering
\includegraphics[width=.99\textwidth]{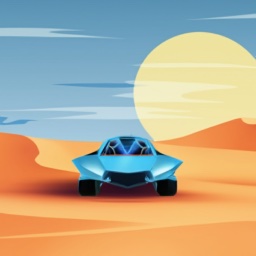}
\caption{\textit{A futuristic car driving through the desert}}
 \end{subfigure} \hfill
\begin{subfigure}[t]{0.33\textwidth}
\centering
\includegraphics[width=.99\textwidth]{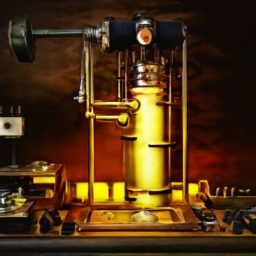}
\caption{\textit{A distillation machine on a table creating gold}}
\end{subfigure} \hfill
\begin{subfigure}[t]{0.33\textwidth}
\includegraphics[width=.99\textwidth]{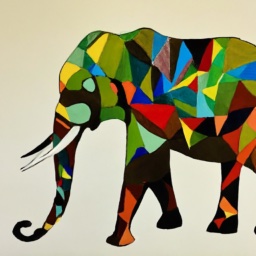}
\caption{\textit{An abstract painting of an elephant}}
\end{subfigure} \hfill
\begin{subfigure}[t]{0.33\textwidth}
\centering
\includegraphics[width=.99\textwidth]{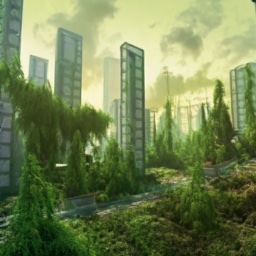}
\caption{\textit{A futuristic city overgrown by nature}}
\end{subfigure} \hfill
\begin{subfigure}[t]{0.33\textwidth}
\centering
\includegraphics[width=.99\textwidth]{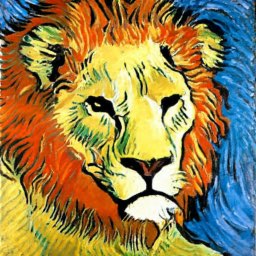}
\caption{\textit{A Van Gogh painting of a lion}}
\end{subfigure}
\begin{subfigure}[t]{0.33\textwidth}
\centering
\includegraphics[width=.99\textwidth]{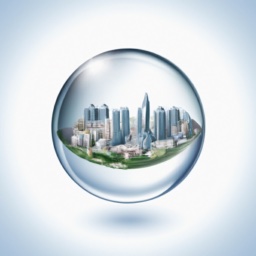}
\caption{\textit{A city inside a glass pearl}}
\end{subfigure} \hfill
\begin{subfigure}[t]{0.33\textwidth}
\centering
\includegraphics[width=.99\textwidth]{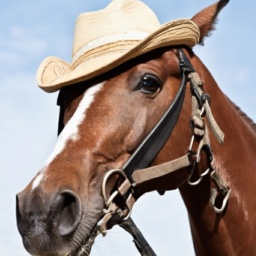}
\caption{\textit{A horse wearing a hat}}
\end{subfigure} \hfill
\begin{subfigure}[t]{0.33\textwidth}
\centering
\includegraphics[width=.99\textwidth]{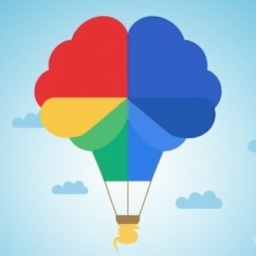}
\caption{\textit{A balloon in the shape of the Google Brain logo}}
\end{subfigure}
\caption{Text to image samples at resolution $256 \times 256$, generated by a single stage diffusion model}
\end{figure*}

\begin{figure}
    \centering
    \includegraphics[width=.3\textwidth]{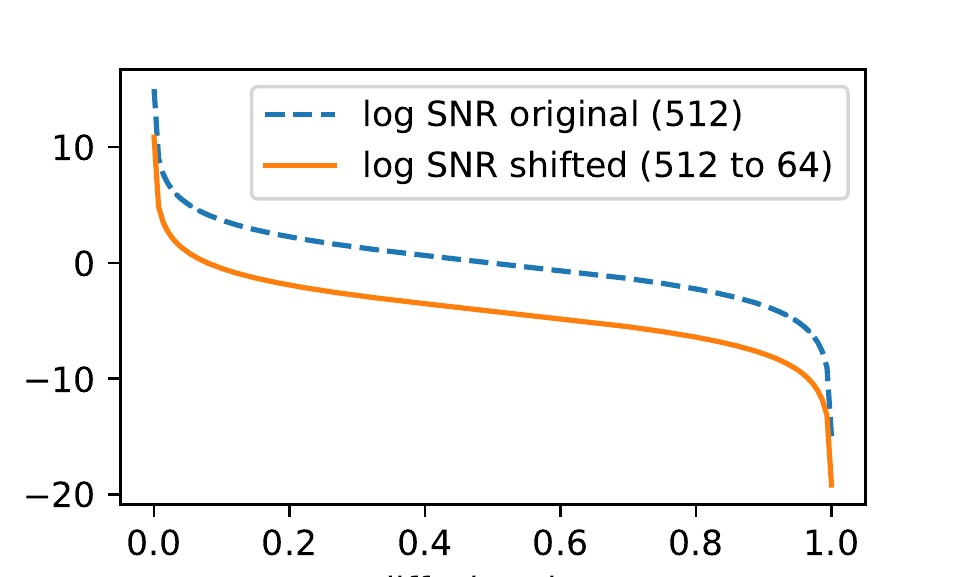}\vspace{-.1cm}
    \caption{Log signal to noise ratio for the original and shifted cosine schedule.}
    \label{fig:diffusion_shifted_snr}
    \vspace{-.35cm}
\end{figure}

One can further derive that the $\alpha_t$ to $\sigma_t$ ratio at this lower resolution is twice as high, meaning that the signal to noise ratio is $2^2$ as high. And so $\mathrm{SNR}^{64 \times 64}(t) = \mathrm{SNR}^{128\times 128}(t) \cdot 2^2$, or in general:
\begin{equation}\small
    \mathrm{SNR}^{d / s \times d / s}(t) = \mathrm{SNR}^{d \times d}(t) \cdot s^2
\end{equation}
In summary, after averaging over a window of size $s \times s$, the ratio $\alpha_t$ to $\sigma_t$ increases by a factor $s$ (and thus the $\mathrm{SNR}$ by $s^2$). Hence, we argue that the noise schedule could be defined with respect to some reference resolution, say $32 \times 32$ or $64 \times 64$ for which the schedules were initially designed and successfully tested. 
In our approach one first chooses a reference resolution, for example $64 \times 64$ (a reasonable choice as we will see empirically). At the reference resolution we define the noise schedule $\mathrm{SNR}^{64 \times 64}(t) = 1 / \tan(\pi t / 2)^2$ which in turn defines the desired $\mathrm{SNR}$ at full resolution $d \times d$:
\begin{equation}
    \mathrm{SNR}_{\mathrm{shift}\, 64}^{d \times d}(t) = \mathrm{SNR}^{64 \times 64}(t) \cdot (64 / d)^2,
\end{equation}
the signal to noise ratio is simply multiplied by $(64 / d)^2$, which for our setting $d > 64$ reduces the signal-to-noise ratio at high resolution. In log-space, this implies a simple shift of $2 \cdot \log (64 / d)$ (see Figure~\ref{fig:diffusion_shifted_snr}). For example, the equation of a noise schedule for images of 128 $\times$ 128 and a reference resolution of 64 the schedule is:
\begin{equation*}
    \log \mathrm{SNR}_{\mathrm{shift} \, 64}^{128 \times 128}(t) = - 2 \log \tan (\pi t / 2) + 2 \log (64 / 128).
\end{equation*}
Recall that under a variance preserving process, the diffusion parameters can be computed as $\alpha^2_t = \mathrm{sigmoid}(\log \mathrm{SNR}(t))$ and $\sigma^2_t = \mathrm{sigmoid}(-\log \mathrm{SNR}(t))$.

Finally, it may be worthwhile to study the concurrent and complementary work \citep{chen2023importancenoise} which also analyzes adjusted noise schedules for higher resolution images and describes several other improvements as well.

\paragraph{Interpolating schedules}
A potential downside of shifting the schedule is that high frequency details are now generated much later in the diffusion process due to the increased per-pixel noise. However, we postulate that high-frequency details are weakly correlated when conditioning on the global/low-frequency features that are already generated. It should therefore be possible to generate the high-frequency details in few diffusion steps. Alternatively, one can \textit{interpolate} different shift schedules, for example for a resolution of $512$ one could include higher frequency details by starting at shift 32 and interpolating in log-space to shift 256. The schedule for $\log \mathrm{SNR}_{\mathrm{interpolate} (32 \to 256)}(t)$ equals:
\begin{equation}\small
     t \log \mathrm{SNR}_{\mathrm{shift} \, 256}^{512 \times 512}(t) + (1 - t) \log \mathrm{SNR}_{\mathrm{shift} \, 32}^{512 \times 512}(t)
\end{equation}
which has more equal weighting over low, mid and high frequency details. When sampling guidance is desired (for example in our text to image experiments) we recommend using this interpolated schedule. We found that shifted schedules can only tolerate little guidance, and interpolated schedules get better results with higher guidance weights.

\subsection{Multiscale training loss}
In the last section we argued that the noise schedule of our diffusion model should be adjusted when training on high resolution images so that the signal-to-noise ratio at our base resolution is held constant. However, even when adjusting the noise schedule in this way, the training loss on images of increasingly high resolution is dominated by high frequency details. To correct for this we propose replacing the standard training loss by a multiscale version that evaluates the standard training loss at downsampled resolutions with a weighting factor that increases for the lower resolutions. We find that the multiscale loss enables quicker convergence especially at resolutions greater than $256 \times 256$. The training loss at the $d \times d$ resolution can be written as:
\[L^{d \times d}_{\theta}(\vx) = \frac{1}{d^{2}}\mathbb{E}_{\veps,t} \lVert \text{D}^{d \times d}[\veps] - \text{D}^{d \times d}[\hat{\veps}_{\theta}(\alpha_t \vx+\sigma_t \veps, t)] \rVert_{2}^{2},\]
where $\text{D}^{d \times d}$ denotes downsampling to the $d \times d$ resolution. If this resolution is identical to the native resolution of our model $\hat{\veps}_{\theta}$ and data $\vx$, the downsampling does not do anything and can be removed from this equation. Otherwise, $\text{D}^{d \times d}[\hat{\veps}_{\theta}]$ can be considered as an adjusted denoising model for data at non-native resolution $d \times d$. Since downsampling an image is a linear operation, we have that $\text{D}^{d \times d}[\mathbb{E}(\veps|\vx)] = \mathbb{E}(\text{D}^{d \times d}[\veps]|\vx)$, and this way of constructing the lower-resolution model is thus indeed consistent with our original model.

We then propose training our high resolution model against the multiscale training loss comprising of multiple resolutions. For instance for the resolutions $32, 64, \ldots, d$ the loss would be: 
$\tilde{L}^{d \times d}_{\theta}(\vx) = \sum_{s \in \{32, 64, 128, \ldots, d\}} \frac{1}{s} L^{s \times s}_{\theta}(\vx)$.

That is, we train against a weighted sum of training losses for resolutions starting at a base resolution (in this case $32 \times 32$) and always including the final resolution of $d \times d$. We find that losses for higher resolution are noisier on average, and we therefore decrease the relative weight of the loss as we increase the resolution.

\subsection{Scaling the Architecture}
Another question is how to scale the architecture. Typical model architectures half the channels each time the resolution is doubled such that the flops per operation is the same but the number of features doubles. The computational intensity (flops / features) also halves each time the resolution doubles. Low computational intensity leads to poor utilization of the accelerator and large activations result in out-of-memory issues. As such, we prefer to scale on the lower resolutions feature maps. Our hypothesis is that mainly scaling on a particular resolution, namely the $16 \times 16$ resolution is sufficient to improve performance within a range of network sizes we consider. Typically, low resolution operations have relatively small feature maps. To illustrate this, consider for example
\begin{equation*} \small
1024 \text{ (batch) } \times 16 \times 16 \times 1024 \text{ (channel)} \cdot 2 \text{ bytes} / \text{dim}
\end{equation*}
costs $0.5$ GB for a feature map whereas for a $256 \times 256$ feature map with $128$ channels, a feature map costs $16$ GB, given they are stored in a 16 bit float format. 

Parameters have a smaller memory footprint: The typical size of a convolutional kernel is $3^2 \times 128^2  \text{ dimensions} \cdot 4 \text{ bytes} / \text{dims} \cdot 5 \text{ replications } = 2.8$MB and $180$MB for $1024$ channels, with $5$ replications for the gradient, optimizer state and exponential moving average. The point is, at a resolution of $16 \times 16$ both the size of feature maps are manageable at $16^2$ and the required space for the parameters is manageable. Summarizing this back-of-the-envelope calculation in Table~\ref{tab:memory_computation} one can see that for the same memory constraint, one can fit $16$GB $/$ $0.7$GB $\approx 23$ layers at $16 \times 16$ versus only $1$ at $256 \times 256$.

\begin{table}
    \centering
    \caption{Memory and compute for a convolutional layer at the typical sizes encountered in diffusion architectures. Using more channels is usually much cheaper at lower resolutions in terms of memory, $B=1024$ for this example.}
    \label{tab:memory_computation}
    \scalebox{.8}{
    \begin{tabular}{l l l}
    \toprule
    Size & ($B \times 256^2 \times 128)$ & ($B \times 16^2 \times 1024)$  \\ \midrule
    Conv Kernel Memory & 2.8MB & 180MB \\
    Feature Map Memory & 16GB & 0.5GB \\ 
    Total Memory & 16GB & 0.7GB \\  \midrule
    Compute (TFLOPS) & 9 & 2.3 \\
        \bottomrule
    \end{tabular}}\vspace{-.4cm}
\end{table}

Other reasons to choose this resolution is because it is the one at which self-attention starts being used in many existing works in the diffusion literature \citep{ho2020denoising,nichol2021improvedddpm}. Furthermore, it is the $16 \times 16$ resolution at which vision transformers for classification can operate successfully \citep{dosovitskiy2021imageisworth}. Although this may not be the ideal way to scale the architecture, we will show empirically that scaling the $16 \times 16$ level works well.

An observant ML practitioner may have realized that when using multiple devices naively, parameters are replicated (typical in JAX and Flax) or stored on the first device (PyTorch). Both cases result in a situation where the memory requirements per device for the feature maps decreases with $1 / \text{devices}$ as desired, but the parameter requirement is unaffected and requires a lot of memory. We scale mostly at a low resolution where activations are relatively small but parameter matrices are large $O(\text{features}^2)$. We found that sharding the weights allows us to scale to much larger models without requiring more complicated parallelization approaches like model parallelism.

\paragraph{Avoiding high resolution feature maps}
High resolution feature maps are memory expensive. If the number of FLOPs is kept constant, memory still scales linearly with the resolution.

In practise, it is not possible to decrease the channels beyond a certain size without sacrificing accelerator utilization. Modern accelerators have a very high ratio between compute and memory bandwidth. Therefore, a low channel count can make operation memory bound, causing a mostly idling accelerator and worse than expected wall-clock performance.

To avoid doing computations on the highest resolutions, we down-sample images immediately as a the first step of the neural network, and up-sample as the last step. Surprisingly, even though the neural networks are cheaper computationally and in terms of memory, we find empirically that they also achieve better performance. We have two approaches to choose from.

\begin{figure}
    \centering
    \includegraphics[width=0.235\textwidth]{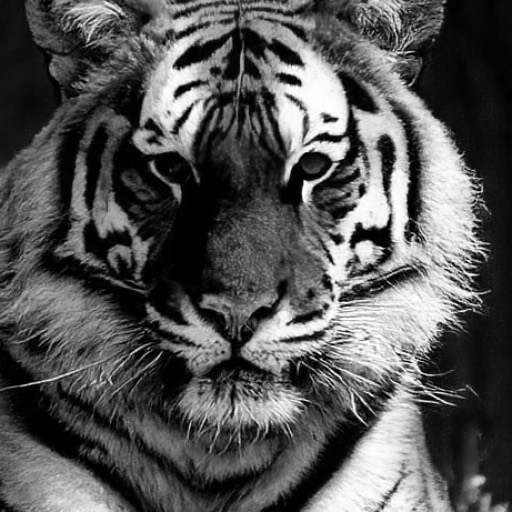}\hfill
    \includegraphics[width=0.235\textwidth]{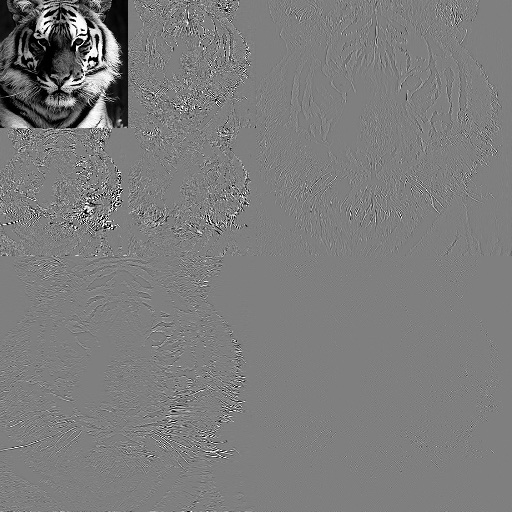}
    \caption{The `5/3' DWT transform transforms an image to low and high frequency responses. Left: original image. Right: The different frequency responses of a two-level DWT, outputs are four $128 \times 128$ maps and three $256 \times 256$ maps. Best viewed electronically.} \vspace{-.6cm}
    \label{fig:dwt}
    \vspace{-.2cm}
\end{figure}

One approach is to use the invertible and linear 5/3 wavelet (as used in JPEG2000) to transform the image to lower resolution frequency responses as demonstrated in Figure~\ref{fig:dwt}. Here, the different feature responses are concatenated spatially for visual purposes. In the network, the responses are concatenated over the channel axis. When more than one level of DWT is applied (here there are two), then the responses differ in resolution. This is resolved by finding the lowest resolution (in the figure $128^2$) and reshaping pixels for the higher resolution feature maps, in the case of $256^2$ they are reshaped $128^2 \times 4$, as a typical space to depth operations. A guide on the implementation of the DWT can be found here\footnote{\url{http://trueharmoniccolours.co.uk/Blog/?p=14}}.

If the above seems to complicated, there also exists a simpler solution if one is willing to pay a small performance penalty. As a first layer one can use a $d \times d$ convolutional layer with stride $d$, and an identically shaped \textit{transposed} convolutional layer as a last layer. This is equivalent to what is called patching in transformer literature. Empirically we show this performs similarly, albeit slightly worse.

\subsection{Dropout}
In architecture typically used in diffusion, a global dropout hyperparameter is used for the residual blocks, at all resolutions. In CDM \citep{ho2022cascaded}, dropout is used to generate images at lower resolutions. For the conditional higher resolution images, no dropout is used. However, various other forms of augmentation are performed on the data. This indicates that regularization is important, even for models operating on high resolutions. However, as we will demonstrate empirically, the naive method of adding dropout in all residual blocks does not give desired results. 

Since our network design only scales the network size at lower resolutions, we hypothesize that it should be sufficient to only add dropout add the lower resolutions. This avoids regularizing the high resolution layers which are memory-wise expensive, while still using the dropout regularization that has been successful for models trained on lower resolution images.

\subsection{The U-ViT architecture}
Taken the above described changes to the architecture one step further, one can replace convolutional layers with MLP blocks if the architecture already uses self-attention at that resolution. This bridges the transformers for diffusion introduced by \citep{peebles2022scalable} with U-Nets, replacing its backbone with a transformer. Consequently, this relatively small change means that we now are using transformer blocks at these resolutions. The main benefit is that the combination of self-attention and MLP blocks has high accelerator utilization, and thus large models train somewhat faster. See Appendix~\ref{app:experimental_details} for details regarding this architecture. In essence, this U-Vision Transformer (U-ViT) architecture can be seen as a small convolutional U-Net which through multiple levels down-samples to the $16 \times 16$ resolution. At this stage a large transformer is applied after which the upsampling is again done via the convolutional U-Net.

\subsection{Text to image generation}
As a proof of concept, we also train a simple diffusion model conditioned on text data. Following \citep{saharia2022imagen} we use the T5 XXL \citep{raffel2020exploring} text encoder as conditioning. For further details see Appendix~\ref{app:experimental_details}. We train three models: One on images of resolution $256 \times 256$ for a direct comparison to models in literature, one on $512 \times 512$ and one on $384 \times 640$. For the last, non-square resolution, images are rotated during prepossessing if their width is smaller than their height, along which a `portrait mode' flag is set to true. As a result, this model can generate natively in a 5:3 aspect ratio for both landscape and portrait orientation.

\section{Related Work}
Score-based diffusion models \citep{sohldickstein2015diffusion,song2019generativemodellingestimatinggradient,ho2020denoising} are a generative model that pre-defines a stochastic destruction process. The generative process is learned by approximating the reverse process with the help of neural networks. 

Diffusion models have been succesfully applied to image generation \citep{ho2020denoising,ho2022cascaded}, speech generation \citep{chen2020wavegrad,kong2021diffwave}, video generation \citep{singer2022makeavideo,saharia2022imagen}. Other types of generative models have also been successfully applied to image generation \citep{chang2022maskgit,sauer2022styleganxl,anonymous2023discrete}, although modifications such as guidance and low temperature sampling can make it difficult to compare these models fairly.
Diffusion models for high resolutions (for example $512^2, 256^2, 128^2$) on complicated data (such as ImageNet) are generally not learned directly. Instead, approaches in literature divide the generative process into sub-problems via super-resolution \citep{ho2022cascaded}, or mixtures-of-denoisers \citep{feng2022ernievilg,balaji2022ediffi}. Alternatively, other approaches project high resolution data down to a lower dimensional latent space \citep{rombach2022highresolution}. Although this sub-division makes optimization easier, the engineering complexity increases: Instead of dealing with a single model, one needs to train and keep track of multiple models. In \citep{gu2022fdm} a different approach to adapt noise to resolution is proposed, although this method seems to generate lower quality samples with a more complicated scheme. We show that it is possible to train a single denoising diffusion model for resolutions up to $512 \times 512$ with only a small number modifications with respect to the original (modern) formulation in \citep{ho2020denoising}.

\section{Experiments}
\label{sec:results}

\subsection{Effects of the proposed modifications}

\begin{table}
    \centering
    \caption{Noise Schedule on ImageNet 128 and 256.}\vspace{-.2cm}
    \label{tab:i128_i256_noise_schedule}
    \scalebox{.9}{
    \begin{tabular}{l l l}
    \toprule
    Noise Schedule & FID train & FID eval \\ \midrule
    \textbf{128 $\times$ 128 resolution} \\
    cosine (original at 128) & 2.96 & 3.38 \\
    cosine (shifted to 64) & 2.41 & 3.03 \\
    cosine (shifted to 32) & \textbf{2.26} & \textbf{2.88} \\ \midrule
    \textbf{256 $\times$ 256 resolution} \\
    cosine (original at 256) & 7.65 & 6.87 \\
    cosine (shifted to 128) & 5.05 & 4.74 \\
    cosine (shifted to 64) & 3.94 & 3.89 \\
    cosine (shifted to 32) & \textbf{3.76} & \textbf{3.71} \\
        \bottomrule
    \end{tabular}}\vspace{-.4cm}
\end{table}

\begin{table}
    \centering
    \caption{Dropout Ablation on ImageNet 128}\vspace{-.2cm}
    \label{tab:dropout}
    \scalebox{.9}{
    \begin{tabular}{l r r r r}
    \toprule
    Starting from Resolution & FID train & FID eval \\ \midrule
    128 & 3.19 & 3.85 \\
    64 & \textbf{2.27} & \textbf{2.85} \\
    32 & 2.31 & 2.87 \\
    16 & 2.41 & 3.03 \\
    no dropout (at 700K iters) & 3.74 & 3.91 \\
        \bottomrule
    \end{tabular}} \vspace{-.4cm}
\end{table}

\textbf{Noise schedule}
In this experiment it is studied how the noise schedule effects the quality of generated images, evaluated on FID50K score on both train and eval data splits. Recall that our hypothesis was that the cosine schedule does not add sufficient noise, but can be adjusted by `shifting' its log SNR curve using the ratio between the image resolution and the noise resolution. In these experiments, the noise resolution is varied from the original image resolution (corresponding to the conventional cosine schedule) all the way down to $32$ by factors of two.

As can be seen in Table~\ref{tab:i128_i256_noise_schedule} for ImageNet at resolution 128 $\times$ 128 and resolution 256 $\times$ 256, shifting the noise schedule considerably improves performance. The difference is especially noticeable at the higher resolution, where the difference is 7.65 for the original cosine schedule against 3.76 for the shifted schedule in FID on the train data. Notice that the difference in performance between the shift towards either 64 and 32 is relatively small, albeit slightly better for the 32 shift. Given that the difference is small and that the shift 64 schedule performed slightly better in early iterations, we generally recommend the shift 64 schedule.

\textbf{Dropout}
The ImageNet dataset has roughly 1 million images. As noted by prior work, it is important to regularize the networks to avoid overfitting \citep{ho2022cascaded,dhariwal2021diffusionbeatgans}. Although dropout has been successfully applied to networks at resolutions of $64 \times 64$, it is often disabled for models operating on high resolutions. In this experiment we enable dropout only on a subset of the network layers: Only for resolutions below the given `starting resolution' hyperparameter. For example, if the starting resolution is $32$, then dropout is applied to modules operating on resolutions $32 \times 32$, $16 \times 16$ and $8 \times 8$.

Recall our hypothesis that it should be sufficient to regularize the modules of the network that operate on the lower resolution feature maps. As presented in Table~\ref{tab:dropout}, this hypothesis holds. For this experiment on images of $128 \times 128$, adding dropout from resolutions $64, 32, 16$ all perform comparatively. Although adding dropout from $16 \times 16$ performed a little worse, we use this setting throughout the remainder of the experiments because it converged faster in early iterations.

The experiment also shows two settings that do not work and should be avoided: either adding no dropout, or adding dropout starting from the same resolution as the data. This may explain why dropout for high resolution diffusion has not been widely used thus far: Typically dropout is set as a global parameter for all feature maps at all resolutions, but this experiment shows that such a regularization is too aggressive.

\textbf{Architecture scaling}
In this section we study the effect of increasing the amount of $16 \times 16$ network modules. In U-Nets, the number of blocks hyperparameter typically refers to the number of blocks on the `down' path. In many implementations, the `up' blocks use one additional block. When the table reads `2 + 3' blocks, that means 2 down blocks and 3 up blocks, which would in literature be referred to as 2 blocks.

Generally, increasing the number of modules improves the performance as can be seen in Table~\ref{tab:scaling}. An interesting exception to this is the eval FID going from $8$ to $12$ blocks, which decreases slightly. We believe that this may indicate that the network should be more strongly regularized as it grows. This effect will later be observed to be amplified for the larger U-ViT architectures.

\begin{table}
    \centering
    \caption{Scaling the U-Net architecture}\vspace{-.2cm}
    \label{tab:scaling}
    \scalebox{.95}{
    \begin{tabular}{l l l r}
    \toprule
    \# blocks at $16 \times 16$ & FID train & FID eval & steps / sec \\ \midrule
    2 \textcolor{gray}{+ 3} & 3.42 & 3.59 & \textbf{114}\% \\
    4 \textcolor{gray}{+ 5} & 2.98 & 3.29 & 100\% \\
    8 \textcolor{gray}{+ 9} & 2.46 & \textbf{3.00} & 76\% \\
    12 \textcolor{gray}{+ 13} & \textbf{2.41} & 3.03 & 62\% \\
        \bottomrule
    \end{tabular}}
\end{table}

\begin{table}
    \centering
    \caption{Downsampling strategies on ImageNet 512 $\times$ 512.}\vspace{-.2cm}
    \label{tab:downsampling}
    \scalebox{.9}{
    \begin{tabular}{l r r r r}
    \toprule
    Strategy & FID train & FID eval & steps / sec \\ \midrule
    None & 5.60 & 5.23 & 100\% \\
    DWT-1 & 5.42 & 4.97 & 139\% \\
    DWT-2 & \textbf{4.85} & \textbf{4.58} & \textbf{146}\% \\
    Conv-(2 $\times$ 2) & 5.99 & 5.33 & 137\% \\
    Conv-(4 $\times$ 4) & 5.04 & 4.80 & \textbf{146}\% \\
        \bottomrule
    \end{tabular}}
    \vspace{-.4cm}
\end{table}

\begin{table}
    \centering
    \caption{Multiscale loss. Note that the $256$ models use the shift 32 and the $512$ use shift 64. This loss modifications is helpful for the highest resolution, but diminishes performance slightly for lower resolutions.}
    \label{tab:multiscale}\scalebox{.9}{
    \begin{tabular}{l l l l}
    \toprule
    Resolution & FID train & FID eval & IS \\ \midrule
    256 & \textbf{3.76} & \textbf{3.71} & \textbf{171.6} \\
    + multiscale loss (32) & 4.00 & 3.89 & 171.0 \\ \midrule
    512 & 4.85 & 4.58 & 156.1 \\
     + multiscale loss (32) & \textbf{4.30} & \textbf{4.28} & \textbf{171.0} \\ 
        \bottomrule
    \end{tabular}}
\end{table}
\begin{table}
    \centering
    \caption{Comparison to generative models in the literature on ImageNet without any guidance or other sampling modifications, except ($^*$) which use temperature scaling.}\vspace{-.2cm}
    \label{tab:literature_comparison}
    \scalebox{.8}{
    \begin{tabular}{l r r l}
    \toprule 
    & \multicolumn{2}{c}{FID} \\
    Method & train & eval & IS \\ \midrule 
    \textbf{128 $\times$ 128 resolution} \\
    ADM \citep{dhariwal2021diffusionbeatgans} & 5.91 \\
    CDM ($32, 64, 128$) \citep{ho2022cascaded} & 3.52 & 3.76 &  128.8 {\small $\pm$ 2.51} \\
    RIN \citep{jabri2022scalable} & 2.75 & & 144.1 \\
    simple diffusion (U-Net) (ours) & 2.26 & \textbf{2.88} & 137.3 {\small $\pm$ 2.03} \\
    simple diffusion (U-ViT 2B) (ours) & \textbf{1.94} & 3.23 & \textbf{171.9} {\small $\pm$ 3.24} \\ \midrule
    \textbf{256 $\times$ 256 resolution} \\
    BigGAN-deep (no truncation) & 6.9\hspace{.175cm} & & 171.4 {\small $\pm$ 2} \\
    MaskGIT \citep{chang2022maskgit} & 6.18 & & 182.1 \\
    DPC$^\star$ (full 5) \citep{anonymous2023discrete} & 4.45 & & \textbf{244.8} \\ \midrule
    \textit{Denoising diffusion models} \\ 
    ADM \citep{dhariwal2021diffusionbeatgans} & 10.94 \\
    CDM ($32, 64, 256$) \citep{ho2022cascaded} & 4.88 & 4.63 & 158.71 {\small $\pm$ 2.26} \\
    LDM-4 \citep{rombach2022highresolution} & 10.56 & & 103.49 \\
    RIN \citep{jabri2022scalable} & 4.51 & & 161.0 \\
    DiT-XL/2 \citep{peebles2022scalable} & 9.62 & & 121.5 \\
    simple diffusion (U-Net) (ours) & 3.76 & \textbf{3.71} & 171.6 {\small $\pm$ 3.07} \\
    simple diffusion (U-ViT 2B) (ours) & \textbf{2.77} & 3.75 & 211.8 {\small $\pm$ 2.93} \\ 
    \midrule
    \textbf{512 $\times$ 512 resolution} \\ 
    MaskGIT \citep{chang2022maskgit} & 7.32 & & 156.0 \\
    DPC (U)$^\star$ \citep{anonymous2023discrete} & 3.62 &  & \textbf{249.4} \\ \midrule
    \textit{Denoising diffusion models} \\ 
    ADM \citep{dhariwal2021diffusionbeatgans}& 23.24 \\
    DiT-XL/2 \citep{peebles2022scalable} & 12.03 & & 105.3 \\
    simple diffusion (U-Net) (ours) & 4.30 & \textbf{4.28} & 171.0 {\small $\pm$ 3.00} \\
    simple diffusion (U-ViT 2B) (ours) & \textbf{3.54} & 4.53 & 205.3{\small $\pm$ 2.65} \\
        \bottomrule
    \end{tabular}}
    \vspace{-.2cm}
\end{table}

\begin{table}
    \centering
    \vspace{-.2cm}
    \caption{Text to image result on zero-shot COCO}\vspace{-.2cm}
    \label{tab:text_to_image_fid}
    \scalebox{.9}{
    \begin{tabular}{l r}
    \toprule
    Method & FID@30K 256 \\ \midrule
    GLIDE \citep{nichol2022glide} & 12.24 \\
    Dalle-2 \citep{ramesh2022hierarchicaltextconditional} & 10.39 \\
    Imagen \citep{saharia2022imagen} & 7.27 \\
    Muse \citep{chang2023muse} & 7.88 \\
    Parti \citep{yu2022scalingautoregressive} & 7.23 \\
    eDiff-I \citep{balaji2022ediffi} & \textbf{6.95} \\
    simple diffusion (U-ViT 2B) (ours) & 8.30 \\
        \bottomrule
    \end{tabular}}
\end{table}

\textbf{Avoiding higher resolution feature maps}
In this experiment, we want to study the effect of downsampling techniques to avoid high resolution feature maps. For this experiment we first have a standard U-Net for images of resolution 512. Then, when we downsample (either to 256 or to 128) using conventional layers or the DWT. For this study the total number of blocks is kept the same, by distributing the high resolution blocks that are skipped over the lower resolution blocks (see Appendix~\ref{app:experimental_details} for more details). Recall our hypothesis that downsampling should not cost much in sample quality, while considerably making the model faster. Surprisingly, in addition to being faster, models that use downsampling strategies also obtain better sample quality. It seems that downsampling for such a high resolution enables the network to optimize better for sample quality. Most importantly, it allows training without absurdly large feature maps without performance degradation.

\textbf{Multiscale Loss}
For this final experiment, we test the difference between the standard loss and the multiscale loss, which adds more emphasis on lower frequencies in the image. For the resolutions 256 and 512 we report the sample quality in FID score for a model trained with the multiscale loss enabled or disabled. As can be seen in Figure~\ref{tab:multiscale}, for 256 the loss does not seem to have much effect and performs slightly worse. However, for the larger 512 resolution the loss has an impact and reduces FID score.

\subsection{Comparison with literature}
In this section, simple diffusion is compared to existing approaches in literature. Although very useful for generating beautiful images, we specifically choose to only compare to methods without guidance (or other sampling modifications such as rejection sampling) to see how well the model is fitted. These sampling modifications may produce inflated scores on visual quality metrics \citep{ho2022classifierfreeguidance}.

Interestingly, the larger U-ViT models perform very well on train FID and Inception Score (IS), outperforming all existing methods in literature (Table~\ref{tab:literature_comparison}). However, the U-Net models perform better on eval FID. We believe this to be an extrapolation of the effect we observed before in Table~\ref{tab:scaling}, where increasing the architecture size did not necessarily result in better eval FID. For samples from the models see Figures~\ref{fig:overview} \& \ref{fig:random_samples_imagenet}.
In summary, simple diffusion achieves SOTA FID scores on class-conditional ImageNet generation among all other types of approaches without sampling modifications. We think this is an incredibly promising result: by adjusting the diffusion schedule and modifying the loss, simple diffusion is a single stage model that operates on resolutions as large as 512 $\times$ 512 with high performance. See Appendix~\ref{app:additional_exps} for additional results.

\textbf{Text to image}
In this experiment we train a text-to-image model following \citep{saharia2022imagen}. In addition to the self-attention and mlp block, this network also has cross-attention in the transformer that operates on T5 XXL text embeddings. For these experiments we also replaced convolutional layers with self-attention at the 32 resolution feature maps to improve detail generation. As can be seen in Table~\ref{tab:text_to_image_fid}, simple diffusion is a little better than some recent text-to-image models such as DALLE-2, although it still lacks behind Imagen. For the resolution $512 \times 512$, the FID@30K score is 9.57. Importantly, our model is the first model that can generate images of this quality using only a single diffusion model that is trained end-to-end.

\section{Conclusion}
In summary, we have introduced several simple modifications of the original denoising diffusion formulation that work well for high resolution images. Without sampling modifiers, simple diffusion achieves state-of-the-art performance on ImageNet in FID score and can be easily trained in an end-to-end setup. Furthermore, to the best of our knowledge this is the first single-stage text to image model that can generate images with such high visual quality.

\clearpage

\bibliography{iclr2022_conference.bib}

\begin{thebibliography}{31}
\providecommand{\natexlab}[1]{#1}
\providecommand{\url}[1]{\texttt{#1}}
\expandafter\ifx\csname urlstyle\endcsname\relax
  \providecommand{\doi}[1]{doi: #1}\else
  \providecommand{\doi}{doi: \begingroup \urlstyle{rm}\Url}\fi

\bibitem[Anonymous(2023)]{anonymous2023discrete}
Anonymous.
\newblock Discrete predictor-corrector diffusion models for image synthesis.
\newblock In \emph{Submitted to The Eleventh International Conference on
  Learning Representations}, 2023.
\newblock URL \url{https://openreview.net/forum?id=VM8batVBWvg}.
\newblock under review.

\bibitem[Balaji et~al.(2022)Balaji, Nah, Huang, Vahdat, Song, Kreis, Aittala,
  Aila, Laine, Catanzaro, Karras, and Liu]{balaji2022ediffi}
Balaji, Y., Nah, S., Huang, X., Vahdat, A., Song, J., Kreis, K., Aittala, M.,
  Aila, T., Laine, S., Catanzaro, B., Karras, T., and Liu, M.
\newblock ediff-i: Text-to-image diffusion models with an ensemble of expert
  denoisers.
\newblock \emph{CoRR}, abs/2211.01324, 2022.

\bibitem[Chang et~al.(2022)Chang, Zhang, Jiang, Liu, and
  Freeman]{chang2022maskgit}
Chang, H., Zhang, H., Jiang, L., Liu, C., and Freeman, W.~T.
\newblock Maskgit: Masked generative image transformer.
\newblock In \emph{{IEEE/CVF} Conference on Computer Vision and Pattern
  Recognition, {CVPR} 2022, New Orleans, LA, USA, June 18-24, 2022}, pp.\
  11305--11315. {IEEE}, 2022.

\bibitem[Chang et~al.(2023)Chang, Zhang, Barber, Maschinot, Lezama, Jiang,
  Yang, Murphy, Freeman, Rubinstein, Li, and Krishnan]{chang2023muse}
Chang, H., Zhang, H., Barber, J., Maschinot, A., Lezama, J., Jiang, L., Yang,
  M., Murphy, K., Freeman, W.~T., Rubinstein, M., Li, Y., and Krishnan, D.
\newblock Muse: Text-to-image generation via masked generative transformers.
\newblock \emph{CoRR}, abs/2301.00704, 2023.

\bibitem[Chen et~al.(2020)Chen, Zhang, Zen, Weiss, Norouzi, and
  Chan]{chen2020wavegrad}
Chen, N., Zhang, Y., Zen, H., Weiss, R.~J., Norouzi, M., and Chan, W.
\newblock {WaveGrad}: {Estimating} gradients for waveform generation.
\newblock \emph{arXiv preprint arXiv:2009.00713}, 2020.

\bibitem[Chen(2023)]{chen2023importancenoise}
Chen, T.
\newblock On the importance of noise scheduling for diffusion models.
\newblock \emph{arxiv}, 2023.

\bibitem[Dhariwal \& Nichol(2021)Dhariwal and
  Nichol]{dhariwal2021diffusionbeatgans}
Dhariwal, P. and Nichol, A.
\newblock Diffusion models beat gans on image synthesis.
\newblock \emph{CoRR}, abs/2105.05233, 2021.

\bibitem[Dosovitskiy et~al.(2021)Dosovitskiy, Beyer, Kolesnikov, Weissenborn,
  Zhai, Unterthiner, Dehghani, Minderer, Heigold, Gelly, Uszkoreit, and
  Houlsby]{dosovitskiy2021imageisworth}
Dosovitskiy, A., Beyer, L., Kolesnikov, A., Weissenborn, D., Zhai, X.,
  Unterthiner, T., Dehghani, M., Minderer, M., Heigold, G., Gelly, S.,
  Uszkoreit, J., and Houlsby, N.
\newblock An image is worth 16x16 words: Transformers for image recognition at
  scale.
\newblock In \emph{9th International Conference on Learning Representations,
  {ICLR} 2021, Virtual Event, Austria, May 3-7, 2021}. OpenReview.net, 2021.

\bibitem[Feng et~al.(2022)Feng, Zhang, Yu, Fang, Li, Chen, Lu, Liu, Yin, Feng,
  Sun, Tian, Wu, and Wang]{feng2022ernievilg}
Feng, Z., Zhang, Z., Yu, X., Fang, Y., Li, L., Chen, X., Lu, Y., Liu, J., Yin,
  W., Feng, S., Sun, Y., Tian, H., Wu, H., and Wang, H.
\newblock Ernie-vilg 2.0: Improving text-to-image diffusion model with
  knowledge-enhanced mixture-of-denoising-experts.
\newblock \emph{CoRR}, abs/2210.15257, 2022.

\bibitem[Gu et~al.(2022)Gu, Zhai, Zhang, Bautista, and Susskind]{gu2022fdm}
Gu, J., Zhai, S., Zhang, Y., Bautista, M.~{\'{A}}., and Susskind, J.~M.
\newblock f-dm: {A} multi-stage diffusion model via progressive signal
  transformation.
\newblock \emph{CoRR}, abs/2210.04955, 2022.

\bibitem[Ho \& Salimans(2022)Ho and Salimans]{ho2022classifierfreeguidance}
Ho, J. and Salimans, T.
\newblock Classifier-free diffusion guidance.
\newblock \emph{CoRR}, abs/2207.12598, 2022.
\newblock \doi{10.48550/arXiv.2207.12598}.
\newblock URL \url{https://doi.org/10.48550/arXiv.2207.12598}.

\bibitem[Ho et~al.(2020)Ho, Jain, and Abbeel]{ho2020denoising}
Ho, J., Jain, A., and Abbeel, P.
\newblock Denoising diffusion probabilistic models.
\newblock In Larochelle, H., Ranzato, M., Hadsell, R., Balcan, M., and Lin, H.
  (eds.), \emph{Advances in Neural Information Processing Systems 33: Annual
  Conference on Neural Information Processing Systems 2020, NeurIPS}, 2020.

\bibitem[Ho et~al.(2022)Ho, Saharia, Chan, Fleet, Norouzi, and
  Salimans]{ho2022cascaded}
Ho, J., Saharia, C., Chan, W., Fleet, D.~J., Norouzi, M., and Salimans, T.
\newblock Cascaded diffusion models for high fidelity image generation.
\newblock \emph{J. Mach. Learn. Res.}, 23:\penalty0 47:1--47:33, 2022.

\bibitem[Jabri et~al.(2022)Jabri, Fleet, and Chen]{jabri2022scalable}
Jabri, A., Fleet, D.~J., and Chen, T.
\newblock Scalable adaptive computation for iterative generation.
\newblock \emph{CoRR}, abs/2212.11972, 2022.

\bibitem[Kingma et~al.(2021)Kingma, Salimans, Poole, and Ho]{kingma2021vdm}
Kingma, D.~P., Salimans, T., Poole, B., and Ho, J.
\newblock Variational diffusion models.
\newblock \emph{CoRR}, abs/2107.00630, 2021.

\bibitem[Kong et~al.(2021)Kong, Ping, Huang, Zhao, and
  Catanzaro]{kong2021diffwave}
Kong, Z., Ping, W., Huang, J., Zhao, K., and Catanzaro, B.
\newblock {DiffWave}: {A} versatile diffusion model for audio synthesis.
\newblock In \emph{9th International Conference on Learning Representations,
  {ICLR}}, 2021.

\bibitem[Meng et~al.(2022)Meng, Gao, Kingma, Ermon, Ho, and
  Salimans]{meng2022ondistillation}
Meng, C., Gao, R., Kingma, D.~P., Ermon, S., Ho, J., and Salimans, T.
\newblock On distillation of guided diffusion models.
\newblock \emph{CoRR}, abs/2210.03142, 2022.

\bibitem[Nichol \& Dhariwal(2021)Nichol and Dhariwal]{nichol2021improvedddpm}
Nichol, A.~Q. and Dhariwal, P.
\newblock Improved denoising diffusion probabilistic models.
\newblock In Meila, M. and Zhang, T. (eds.), \emph{Proceedings of the 38th
  International Conference on Machine Learning, {ICML}}, 2021.

\bibitem[Nichol et~al.(2022)Nichol, Dhariwal, Ramesh, Shyam, Mishkin, McGrew,
  Sutskever, and Chen]{nichol2022glide}
Nichol, A.~Q., Dhariwal, P., Ramesh, A., Shyam, P., Mishkin, P., McGrew, B.,
  Sutskever, I., and Chen, M.
\newblock {GLIDE:} towards photorealistic image generation and editing with
  text-guided diffusion models.
\newblock In Chaudhuri, K., Jegelka, S., Song, L., Szepesv{\'{a}}ri, C., Niu,
  G., and Sabato, S. (eds.), \emph{International Conference on Machine
  Learning, {ICML} 2022, 17-23 July 2022, Baltimore, Maryland, {USA}}, volume
  162 of \emph{Proceedings of Machine Learning Research}, pp.\  16784--16804.
  {PMLR}, 2022.
\newblock URL \url{https://proceedings.mlr.press/v162/nichol22a.html}.

\bibitem[Peebles \& Xie(2022)Peebles and Xie]{peebles2022scalable}
Peebles, W. and Xie, S.
\newblock Scalable diffusion models with transformers.
\newblock \emph{CoRR}, abs/2212.09748, 2022.

\bibitem[Raffel et~al.(2020)Raffel, Shazeer, Roberts, Lee, Narang, Matena,
  Zhou, Li, and Liu]{raffel2020exploring}
Raffel, C., Shazeer, N., Roberts, A., Lee, K., Narang, S., Matena, M., Zhou,
  Y., Li, W., and Liu, P.~J.
\newblock Exploring the limits of transfer learning with a unified text-to-text
  transformer.
\newblock \emph{J. Mach. Learn. Res.}, 21:\penalty0 140:1--140:67, 2020.

\bibitem[Ramesh et~al.(2022)Ramesh, Dhariwal, Nichol, Chu, and
  Chen]{ramesh2022hierarchicaltextconditional}
Ramesh, A., Dhariwal, P., Nichol, A., Chu, C., and Chen, M.
\newblock Hierarchical text-conditional image generation with {CLIP} latents.
\newblock \emph{CoRR}, abs/2204.06125, 2022.

\bibitem[Rombach et~al.(2022)Rombach, Blattmann, Lorenz, Esser, and
  Ommer]{rombach2022highresolution}
Rombach, R., Blattmann, A., Lorenz, D., Esser, P., and Ommer, B.
\newblock High-resolution image synthesis with latent diffusion models.
\newblock In \emph{{IEEE/CVF} Conference on Computer Vision and Pattern
  Recognition, {CVPR} 2022, New Orleans, LA, USA, June 18-24, 2022}, pp.\
  10674--10685. {IEEE}, 2022.

\bibitem[Saharia et~al.(2022)Saharia, Chan, Saxena, Li, Whang, Denton,
  Ghasemipour, Ayan, Mahdavi, Lopes, Salimans, Ho, Fleet, and
  Norouzi]{saharia2022imagen}
Saharia, C., Chan, W., Saxena, S., Li, L., Whang, J., Denton, E., Ghasemipour,
  S. K.~S., Ayan, B.~K., Mahdavi, S.~S., Lopes, R.~G., Salimans, T., Ho, J.,
  Fleet, D.~J., and Norouzi, M.
\newblock Photorealistic text-to-image diffusion models with deep language
  understanding.
\newblock \emph{CoRR}, abs/2205.11487, 2022.

\bibitem[Salimans \& Ho(2022)Salimans and Ho]{salimans2022progressive}
Salimans, T. and Ho, J.
\newblock Progressive distillation for fast sampling of diffusion models.
\newblock In \emph{The Tenth International Conference on Learning
  Representations, {ICLR}}. OpenReview.net, 2022.

\bibitem[Sauer et~al.(2022)Sauer, Schwarz, and Geiger]{sauer2022styleganxl}
Sauer, A., Schwarz, K., and Geiger, A.
\newblock Stylegan-xl: Scaling stylegan to large diverse datasets.
\newblock In Nandigjav, M., Mitra, N.~J., and Hertzmann, A. (eds.),
  \emph{{SIGGRAPH} '22: Special Interest Group on Computer Graphics and
  Interactive Techniques Conference}, pp.\  49:1--49:10. {ACM}, 2022.

\bibitem[Singer et~al.(2022)Singer, Polyak, Hayes, Yin, An, Zhang, Hu, Yang,
  Ashual, Gafni, Parikh, Gupta, and Taigman]{singer2022makeavideo}
Singer, U., Polyak, A., Hayes, T., Yin, X., An, J., Zhang, S., Hu, Q., Yang,
  H., Ashual, O., Gafni, O., Parikh, D., Gupta, S., and Taigman, Y.
\newblock Make-a-video: Text-to-video generation without text-video data.
\newblock \emph{CoRR}, abs/2209.14792, 2022.

\bibitem[Sohl{-}Dickstein et~al.(2015)Sohl{-}Dickstein, Weiss, Maheswaranathan,
  and Ganguli]{sohldickstein2015diffusion}
Sohl{-}Dickstein, J., Weiss, E.~A., Maheswaranathan, N., and Ganguli, S.
\newblock Deep unsupervised learning using nonequilibrium thermodynamics.
\newblock In Bach, F.~R. and Blei, D.~M. (eds.), \emph{Proceedings of the 32nd
  International Conference on Machine Learning, {ICML}}, 2015.

\bibitem[Song \& Ermon(2019)Song and
  Ermon]{song2019generativemodellingestimatinggradient}
Song, Y. and Ermon, S.
\newblock Generative modeling by estimating gradients of the data distribution.
\newblock In \emph{Advances in Neural Information Processing Systems 32: Annual
  Conference on Neural Information Processing Systems 2019, NeurIPS}, 2019.

\bibitem[Song et~al.(2021)Song, Sohl{-}Dickstein, Kingma, Kumar, Ermon, and
  Poole]{song2021scorebasedsde}
Song, Y., Sohl{-}Dickstein, J., Kingma, D.~P., Kumar, A., Ermon, S., and Poole,
  B.
\newblock Score-based generative modeling through stochastic differential
  equations.
\newblock In \emph{9th International Conference on Learning Representations,
  {ICLR} 2021, Virtual Event, Austria, May 3-7, 2021}. OpenReview.net, 2021.

\bibitem[Yu et~al.(2022)Yu, Xu, Koh, Luong, Baid, Wang, Vasudevan, Ku, Yang,
  Ayan, Hutchinson, Han, Parekh, Li, Zhang, Baldridge, and
  Wu]{yu2022scalingautoregressive}
Yu, J., Xu, Y., Koh, J.~Y., Luong, T., Baid, G., Wang, Z., Vasudevan, V., Ku,
  A., Yang, Y., Ayan, B.~K., Hutchinson, B., Han, W., Parekh, Z., Li, X.,
  Zhang, H., Baldridge, J., and Wu, Y.
\newblock Scaling autoregressive models for content-rich text-to-image
  generation.
\newblock \emph{CoRR}, abs/2206.10789, 2022.

\end{thebibliography}
\bibliographystyle{icml2022}

\clearpage
\appendix
\onecolumn
\section{Additional Background Information on Diffusion Models}
\label{sec:addition_info_diffusion}
This section is a more detailed summary of relevant background information on denoising diffusion. For one, it can be helpful to understand how modern denoising diffusion models \citep{ho2020denoising} are trained using the formulations from \citep{kingma2021vdm}
First we define how signal is destroyed (diffused), which is the algorithmic equivalent to sampling $\vz_t \sim q(\vz_t | \vx)$:
\begin{lstlisting}[style=python]
def diffuse(x, alpha_t, sigma_t):
  eps_t = noise_normal_like(x)
  z_t = alpha_t * x + sigma_t * eps_t
  return z_t, eps_t
\end{lstlisting}

For the specific optimization setting we generally use (v-prediction, epsilon loss) the loss can be computed as defined below. This is the algorithmic equivalent of $\mathbb{E}_{t \sim \mathcal{U}(0, 1), \vz_t \sim q(\vz_t | \vx)}||f(\vz_t, t) - \eps_t||^2$ as proposed by \citep{ho2020denoising,kingma2021vdm}:
\begin{lstlisting}[style=python]
def loss(x):
  t = noise_uniform(size=x.shape[0]) # Sample a batch of timesteps.
  logsnr_t = logsnr_schedule(t)
  alpha_t = sqrt(sigmoid(logsnr))
  sigma_t = sqrt(sigmoid(-logsnr))
  z_t, eps_t = diffuse(x, alpha_t, sigma_t)
  v_pred = uvit(z_t, logsnr_t)
  eps_pred = sigma_t * z_t + alpha_t * v_t
  return mse(eps_pred, eps_t)
\end{lstlisting}

In case of conditioning (for example ImageNet class number of a text embedding), these are added as an input to the uvit call, but do not influence the diffusion process in other ways. The conditioning is dropped out $10\%$ of the time, so that the models can additionally be used with classifier-free guidance.

The standard cosine logsnr schedule (taking care of boundaries) can be defined as:
\begin{lstlisting}[style=python]
def logsnr_schedule_cosine(t, logsnr_min=-15, logsnr_max=+15):
  t_min = atan(exp(-0.5 * logsnr_max))
  t_max = atan(exp(-0.5 * logsnr_min))
  return -2 * log(tan(t_min + t * (t_max - t_min)))
\end{lstlisting}

One can then define the shifted schedule as:
\begin{lstlisting}[style=python]
def logsnr_schedule_cosine_shifted(t, image_d, noise_d):
  return logsnr_schedule_cosine(t) + 2 log(noise_d / image_d)
\end{lstlisting}

And the interpolated schedule as:
\begin{lstlisting}[style=python]
def logsnr_schedule_cosine_shifted(t, image_d, noise_d_low, noise_d_high):
  logsnr_low = logsnr_schedule_cosine_shifted(t, image_d, noise_d_low)
  logsnr_high = logsnr_schedule_cosine_shifted(t, image_d, noise_d_high)
  return t * logsnr_low + (1 - t) * logsnr_high
\end{lstlisting}

Care needs to be taken that the minimum and maximum logsnr hyperparameters are shifted along with the entire schedule, so care needs to be taken when these endpoints are used to define the embedding in the architecture.

\paragraph{Sampling}
In this work we use the standard ddpm sampler unless noted otherwise. Below is the algorithmic equivalent of the generative process of sampling $\vz_T \sim \mathcal{N}(0, \mathbf{I})$ and then repeatedly sampling $\vz_{s} \sim p(\vz_s | \vz_t)$:

\begin{lstlisting}[style=python]
def sample(x_shape):
  # lowest_idx can be 0 or 1.
  z_t = noise_normal(x_shape)
  for t in reversed(range(lowest_idx+1, num_steps+1)):
    u_t = t / num_steps
    u_s = (t - 1) / num_steps
    logsnr_t = logsnr_schedule(u_t)
    logsnr_s = logsnr_schedule(u_s)
    v_pred = uvit(z_t, logsnr_t)
    z_t = sampler_step(z_t, v_pred, logsnr_t, logsnr_s)

  # Final prediction, do not sample x ~ p(x | z_lowest) but take the mean prediction:
  logsnr_lowest = logsnr_schedule(lowest_idx / num_steps)
  v_pred = uvit(z_t, logsnr_lowest)
  x_pred = alpha_t * z_t - sigma_t * v_pred
  x_pred = clip_x(x_pred)
  return x_pred
\end{lstlisting}

\begin{lstlisting}[style=python]
def ddpm_sampler_step(z_t, v_pred, logsnr_t, logsnr_s):
  x_pred = alpha_t * z_t - sigma_t * v_pred
  x_pred = clip_x(x_pred)
  
  mu = exp(logsnr_t - logsnr_s) * alpha_st * z_t + (1 - exp(logsnr_t - logsnr_s)) * alpha_s * x_pred
  # Variance can be any interpolation of the following two in log-space:
  min_lvar = (1 - exp(logsnr_t - logsnr_s)) + log_sigmoid(-logsnr_s)
  max_lvar = (1 - exp(logsnr_t - logsnr_s)) + log_sigmoid(-logsnr_t)
  noise_param = 0.2
  sigma = sqrt(exp(noise_param * max_logvar + (1 - noise_param) * min_logvar))
  return mu + sigma * normal_noise_like(z_t)
\end{lstlisting}
where noise\_param is set to 0.2 with the exception of MSCOCO FID evaluation, where it is set to 1.0.

An important but not often discussed detail is that during sampling it is helpful to clip the predictions in x-space, below gives an example for static clipping, for dynamic clipping see \citep{saharia2022imagen}:
\begin{lstlisting}[style=python]
def clip_x(x):
  # x should be between -1 and 1.
  return clip(x, -1, 1)
\end{lstlisting}

\paragraph{Classifier-free guidance}
In classifier-free guidance \citep{ho2022classifierfreeguidance}, one drops out the conditioning signal occasionally during training (Usually about 10\% of the time). This allows one to train models, $p(\vx)$ in addition to the model one normally trains which is $p(\vx | \text{cond})$. The epsilon predictions of these models can then be recombined with a guidance scale. For $\eta > 0$:
\begin{equation}
    \hat{\veps}(\vx) = (1 + \eta) \hat{\veps}(\vx , \text{cond}) - \eta \hat{\veps}(\vx).
\end{equation}
One can substitute $\hat{\veps}$ by $\hat{\vv}$ or $\hat{\vx}$ and the result ends up being equivalent due to linearity and terms cancelling out. Note we will report the guidance scale as $(1 + \eta)$ as is done often in literature, not to be confused by reporting $\eta$ itself.

\paragraph{Distillation}
Like many diffusion models, simple diffusion can also be distilled to reduce the number of sampling steps and neural net evaluations \citep{meng2022ondistillation} to reduce the number of sampling steps. For a distilled U-ViT model, generating a single image takes 0.42 seconds on a TPUv4. Similarly, generating a batch of 8 images takes 2.00 seconds.

\section{Experimental details}
\label{app:experimental_details}

In this section, specific details on the experiments are given. Firstly, the standard optimizer settings for the U-Net experiments.
\subsection{U-Net settings}
\begin{lstlisting}
unet default optimization settings:
    batch_size=512,
    optimizer='adam',
    adam_beta1=0.9,
    adam_beta2=0.99,  except for ImageNet 128 which is adam_beta2=0.999
    adam_eps=1.e-12,  
    learning_rate=5e-5,
    learning_rate_warmup_steps=10_000,
    weight_decay=0.0,
    ema_decay=0.9999,
    grad_clip=1.0,
\end{lstlisting}

\begin{lstlisting}
Specific settings for the UNet on ImageNet 128 experiment:
    base_channels=128,
    emb_channels=1024,                              (for diffusion time, image class)
    channel_multiplier=[1, 2, 4, 8, 8],
    num_res_blocks=[3, 4, 4, 12, 4],                (unless noted otherwise)
    attn_resolutions=[8, 16],
    num_heads=4,
    dropout_from_resolution=16,                     (unless noted otherwise)
    dropout=0.1,
    patching_type='none'
    schedule={'name': 'cosine_shifted, 'shift': 64} (unless noted otherwise)
    num_train_steps=1_500_000
\end{lstlisting}

\begin{lstlisting}
Specific settings for the UNet on ImageNet 256 experiment:
    base_channels=128,
    emb_channels=1024,                              (for diffusion time, image class)
    channel_multiplier=[1, 1, 2, 4, 8, 8],
    num_res_blocks=[1, 2, 2, 4, 12, 4],
    attn_resolutions=[8, 16],
    num_heads=4,
    dropout_from_resolution=16,
    dropout=0.1,
    patching_type='none'
    schedule={'name': 'cosine_shifted, 'shift': 64} (unless noted otherwise)
    num_train_steps=2_000_000
\end{lstlisting}

\begin{lstlisting}
Setting for the UNet on ImageNet 512 experiment:
    base_channels=128,
    emb_channels=1024,                              (for diffusion time, image class)
    attn_resolutions=[8, 16],
    num_heads=4,
    dropout_from_resolution=16,
    dropout=0.1,
    patching_type='dwt_2'                           (unless noted otherwise)
    schedule={'name': 'cosine_shifted, 'shift': 64} (unless noted otherwise)
    num_train_steps=2_000_000
\end{lstlisting}
To keep the number of residual blocks the same, high resolution blocks that are skipped by down-sampling are added to the lower resolution levels. With no downsampling, the architecture uses:
\begin{lstlisting}
channel_multiplier=[1, 1, 1, 2, 4, 8, 8], num_res_blocks=[1, 1, 2, 2, 4, 12, 4],
\end{lstlisting}
In case of $2\times$ downsampling the architecture uses:
\begin{lstlisting}
channel_multiplier=[1, 2, 2, 4, 8, 8], num_res_blocks=[2, 2, 2, 4, 12, 4],
\end{lstlisting}
In case of $4\times$ downsampling the architecture uses:
\begin{lstlisting}
channel_multiplier=[2, 3, 4, 8, 8], num_res_blocks=[3, 3, 4, 12, 4],
\end{lstlisting}

\begin{figure}
    \centering
    \includegraphics[width=.8\textwidth]{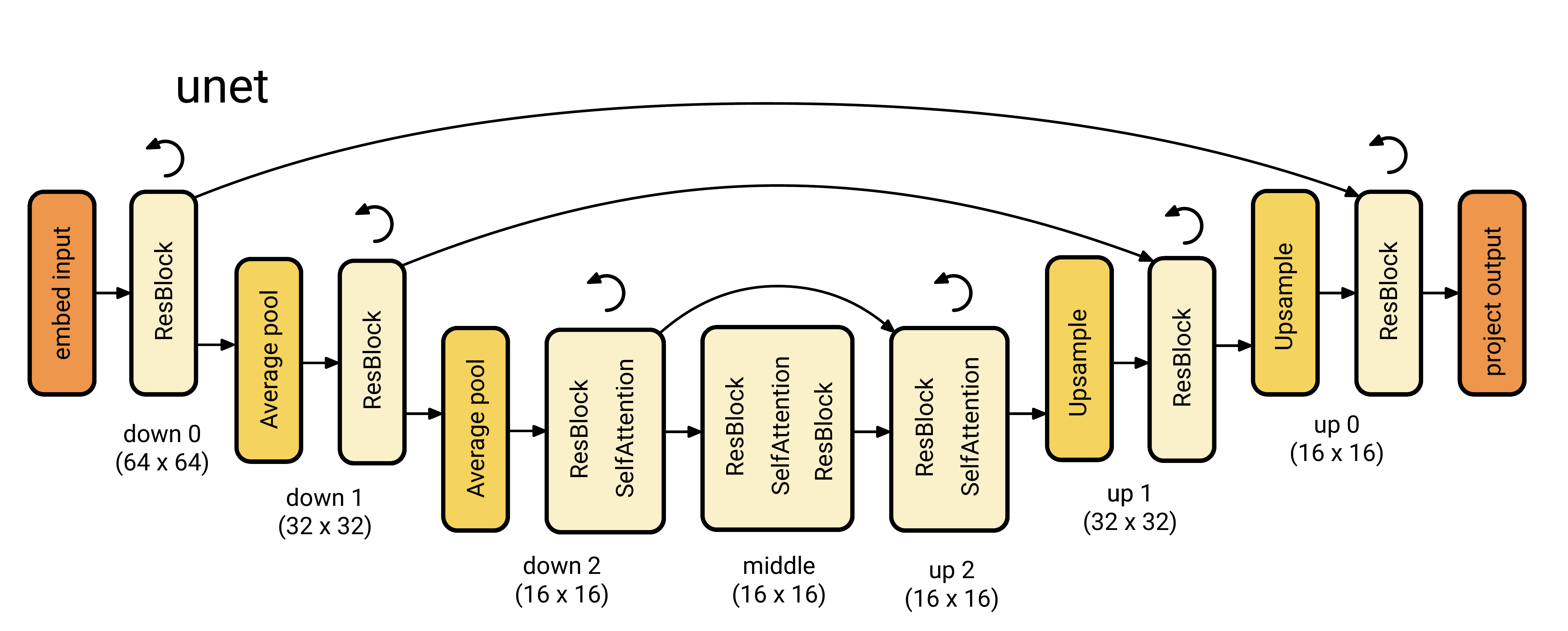} \vspace{.05cm}
    \includegraphics[width=.8\textwidth]{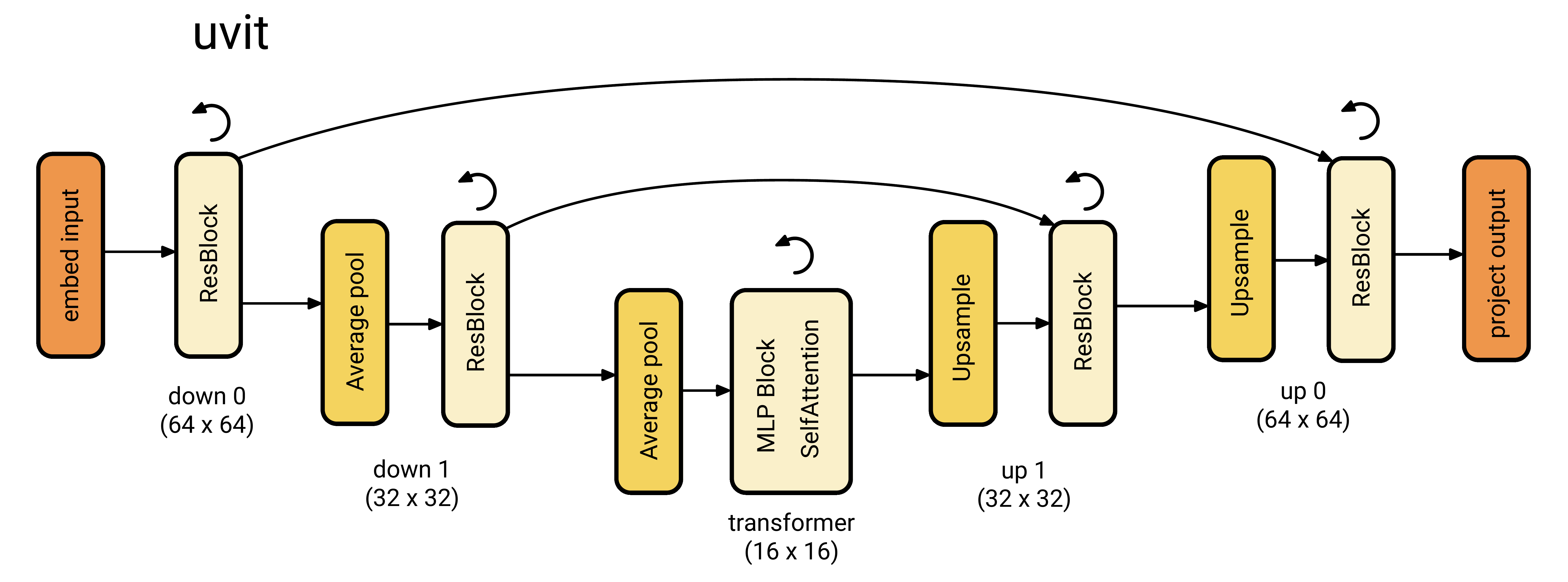}
    \caption{The difference between the U-Net and U-ViT architecture. In essence, the convolutional layers are replaced by MLP blocks on levels with self-attention. These now form transformer blocks which are connected via residual connections, only the ResBlocks on higher levels use skip connections. Circular arrows denote that such a block can be repeated multiple times.}
    \label{fig:unet_vs_uvit}
\end{figure}

\subsection{U-ViT settings}
The U-ViT is a very similar architecture to the U-Net (see Figure~\ref{fig:unet_vs_uvit}). The two major differences are that 1) When a module has self-attention, it uses an MLP block instead of a convolutional layer, making their combination a transformer block. And 2) the transformer blocks in the middle do not use skip connections, only residual connections. The default optimization settings for ImageNet for the U-ViT are:
\begin{lstlisting}
uvit default optimization settings:
    optimizer='adam',
    adam_beta1=0.9,
    adam_beta2=0.99,
    adam_eps=1.e-12,
    learning_rate=1e-4,
    learning_rate_warmup_steps=10_000,
    weight_decay=0.0,
    ema_decay=0.9999,
    grad_clip=1.0,
    batch_size=2048,
    num_train_steps=500_000,
\end{lstlisting}

And the architecture settings are almost the same for all resolutions $128$, $256$ and $512$.
\begin{lstlisting}
uvit default architecture settings for 512:
    optimizer='adam',
    adam_beta1=0.9,
    adam_beta2=0.99,
    adam_eps=1.e-12,
    learning_rate=1e-4,
    learning_rate_warmup_steps=10_000,
    weight_decay=0.0,
    ema_decay=0.9999,
    grad_clip=1.0,
    batch_size=2048,
    base_channels=128,
    emb_channels=1024,
    channel_multiplier=[1, 2, 4, 16],
    num_res_blocks=[2, 2, 2],
    num_transformer_blocks=36,
    num_heads=4,
    transformer_dropout=0.2,
    logsnr_input_type='linear',
    patching_type='dwt_5/3_2',
    mean_type='v',
    mean_loss_type='v_mse',
\end{lstlisting}
where the patching type is either \texttt{'none'} for $128$, \texttt{'dwt\_1'} for 256 and  \texttt{'dwt\_2'} for 512. Note also that the loss is computed on v instead of epsilon. This may not be very important: in small experiments we observed only minor performance differences between the two. Note also that the batch size is larger (2048) which does affect FID and IS performance considerably. The text to image model was trained for 700K steps.

\subsubsection{Pseudo-code for U-ViT modules}
The Transformer blocks consist of a self-attention and mlp block. These are defined as one would expect, for completeness given below in pseudo-code:
\begin{lstlisting}[style=python]
def mlp_block(x, emb, expansion_factor=4):
  B, HW, C = x.shape 
  x = Normalize(x)
  mlp_h = Dense(x, expansion_factor * C)
  scale = DenseGeneral(emb, mlp_h.shape[2:])
  shift = DenseGeneral(emb, mlp_h.shape[2:])

  mlp_h = swish(mlp_h)
  mlp_h = mlp_h * (1. + scale[:, None]) + shift[:, None]
  if config.transformer_dropout > 0.:
    mlp_h = Dropout(mlp_h, config.transformer_dropout)

  out = Dense(mlp_h, C, kernel_init=zeros)
  return out
  
def self_attention(x, text_emb):
  B, HW, C = x.shape
  B, T, TC = text_emb.shape
  head_dim = C // config.num_heads

  x_norm = Normalize(x)

  q = DenseGeneral(x_norm, (num_heads, head_dim))
  k = DenseGeneral(x_norm, (num_heads, head_dim))
  v = DenseGeneral(x_norm, (num_heads, head_dim))

  q = NormalizeWithBias(q)
  k = NormalizeWithBias(k)
  q = q * q.shape[-1] ** -0.5
  weights = einsum("bqhd,bkhd->bhqk", q, k)
  weights = softmax(weights)
  attn_vals = einsum("bhqk,bkhd->bqhd", weights, v)

  out = DenseGeneral(attn_vals, C, axis=(-2, -1), kernel_init=zeros)
  return out
  
def transformer_block(x, text_emb, emb):
  x += mlp_block(x, emb)
  x += self_attention(x, text_emb)
  return x
\end{lstlisting}

Another important block is the standard ResBlock, pseudo-code given below:
\begin{lstlisting}[style=python]
def resnet_block(x, emb, skip_h=None):
  B, H, W, C = x.shape
  h = NormalizeWithBias(x)
  if skip_h is not None:
    skip_h = NormalizeWithBias(skip_h)
    h = (h + skip_h) / sqrt(2)
  h = swish(h)
  h = Conv2D(h, out_ch, (3, 3), (1, 1))
  emb_out = Dense(emb, 2 * out_ch)[:, None, None, :]

  scale, shift = split(emb_out, 2, axis=-1)
  h = NormalizeWithBias(h) * (1 + scale) + shift
  h = swish(h)
  h = Conv2D(h, out_ch, (3, 3), (1, 1), kernel_init=zeros)
  return x + h
\end{lstlisting}

Given these building blocks, one can define the U-ViT architecture:
\begin{lstlisting}[style=python]
def uvit(x, logsnr):
  B, H, W, C = x.shape
  emb = get_logsnr_emb(logsnr)

  h0 = EmbedInput(config.base_channels * config.channel_multiplier[0])(x)
  hs = []
  last_h = h0
  # Down path.
  for i_level in range(len(config.num_res_blocks))):
    for i_block in range(config.num_res_blocks[i_level]):
      last_h = resnet_block(last_h, emb)
      hs.append(last_h)

    last_h = downsample(
        last_h, config.base_channels * config.channel_multiplier[i_level+1])
  
  # The transformer.
  last_h = last_h.reshape(B, H * W, C)
  last_h += param("pos_emb", initializers.normal(0.01), last_h.shape[1:])[None]
  for _ in range(config.num_transformer_blocks):
    last_h = transformer_block(last_h, text_emb, emb)
  last_h = last_h.reshape(B, H, W, C)

  # Up path.
  for i_level in reversed(range(len(config.num_res_blocks)))):
    last_h = upsample(last_h, config.base_channels * config.channel_multiplier[i_level])
    for i_block in range(config.num_res_blocks[i_level]):
      last_h = resnet_block(last_h, emb, skip_h=hs.pop())
  
  out = ProjectOutput(last_h, C)
  return out
\end{lstlisting}

As one can see, it's very similar to the UNet, the middle part is now a transformer which does not have convolutional layers but mlp blocks with only residual connections.

\paragraph{Computational resources}
The smaller U-Net models can be trained on 64 TPUv2 devices with 1.15 steps per second (for a resolution of 256 without patching, small differences between different model variants) with a batch size of 512 for 2000K steps (unless specified otherwise). The large U-ViT models are all trained using 128 TPUv4 devices with 1.5 steps per second with a batch size of 2048 for 500K steps.

\newpage\section{Additional Experiments}
\label{app:additional_exps}

\begin{figure}[H]
    \centering
    \includegraphics[width=0.5\textwidth]{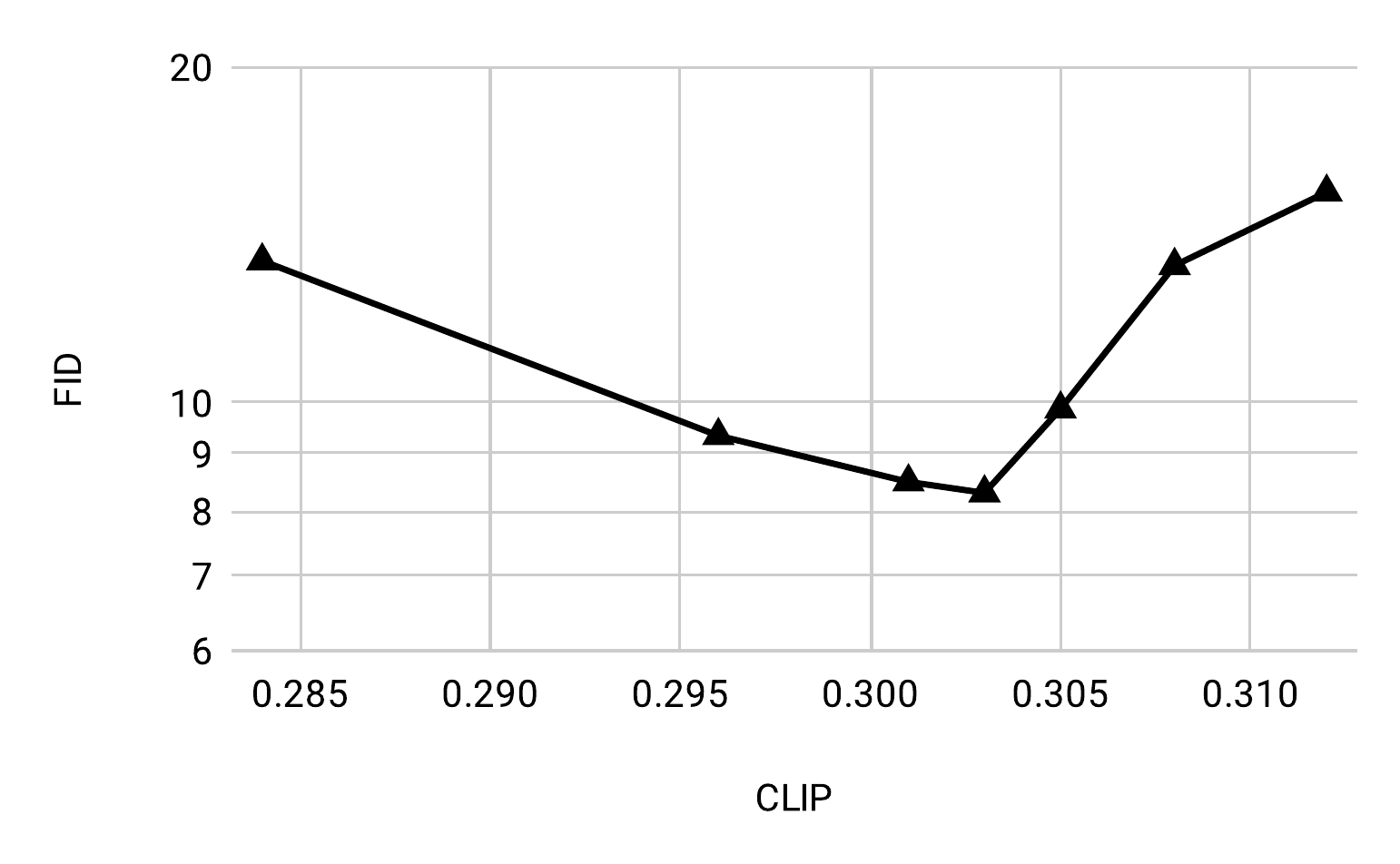}
    \caption{Clip vs FID30K score on zero-shot MSCOCO at resolution 256 $\times$ 256. For guidance scales 1.00, 1.25, 1.40, 1.50, 2.0, 3.0, 4.0.}
    \label{fig:clipvsfid}
\end{figure}

\paragraph{Guidance scale} In Table~\ref{tab:guidance} we show the effect of guidance on the ImageNet models. For relatively small levels of guidance, samples immediately gain a lot in IS at the cost of especially eval FID. Furthermore, Figure~\ref{fig:clipvsfid} shows the Clip versus MSCOCO FID30K score for the text to image model. Following others such as \citep{saharia2022imagen}, images are sampled by conditioning on 30K randomly sampled texts from the MSCOCO validation set, computed against the full validation set as a reference.
\begin{table}[H]
    \centering
    \caption{Guidance scale, the shifted schedule is quite sensitive to guidance.}
    \label{tab:guidance}
    \scalebox{.9}{
    \begin{tabular}{l | r r l | r r l | r r l }
    \toprule
    U-ViT & \multicolumn{3}{c}{ImageNet 128} & \multicolumn{3}{c}{ImageNet 256} & \multicolumn{3}{c}{ImageNet 512} \\ \midrule
    guidance & FID train & FID eval & IS & FID train & FID eval & IS & FID train & FID eval & IS \\ \midrule
    1.00 & \textbf{1.94} & \textbf{3.23} & 171.9 {\small $\pm$ 3.2} & 2.77 & \textbf{3.75} & 211.8 {\small $\pm$ 2.9} & 3.54 & 4.53 & 205.3{\small $\pm$ 2.7} \\
    1.05 & 2.05 & 3.57 & 189.9{\small $\pm$ 3.5} & 2.46 & 3.80 & 235.3 {\small $\pm$ 4.9} & 3.14 & \textbf{4.43} & 228.5{\small $\pm$ 4.2} \\
    1.10 & 2.35 & 4.10 & 207.0{\small $\pm$ 3.5} & \textbf{2.44} & 4.08 & 256.3 {\small $\pm$ 5.0} & \textbf{3.02} & 4.60 & 248.7{\small $\pm$ 3.4} \\
    1.20 & 3.24 & 5.36 & 237.6{\small $\pm$ 3.6} & 2.96 & 5.10 & 289.8 {\small $\pm$ 4.1} & 3.33 & 5.43 & 284.6{\small $\pm$ 2.8} \\
    1.40 & 5.58 & 8.26 & 285.2{\small $\pm$ 2.0} & 4.69 & 7.50 & 342.2 {\small $\pm$ 5.1} & 4.97 & 7.89 & 339.9{\small $\pm$ 3.8} \\
    1.80 & 9.77 & 13.06 & 340.1{\small $\pm$ 3.6} & 8.21 & 11.81 & 398.0 {\small $\pm$ 5.4} & 8.38 & 12.15 & 401.7{\small $\pm$ 5.2} \\
    2.00 & 11.47 & 14.96 & 359.2{\small $\pm$ 5.6} & 9.59 & 13.44 & 416.4 {\small $\pm$ 4.7} & 9.68 & 13.68 & 416.2{\small $\pm$ 4.8} \\
    3.00 & 15.85 & 19.75 & \textbf{399.2}{\small $\pm$ 2.9} & 13.61 & 18.00 & \textbf{455.7} {\small $\pm$ 4.2} & 13.79 & 18.42 & \textbf{461.4}{\small $\pm$ 5.0} \\
        \bottomrule
    \end{tabular}}\vspace{-.2cm}
\end{table}

\paragraph{Experiments on 1024}
To study the effects of scaling beyond 512, we run a similar experiment with U-Nets on ImageNet resized to 1024 by 1024, even though most images are smaller than that resolution. Here, the multiscale loss has an even more pronounced effect, resulting in a train FID that is considerably improved by using the downsample loss (6.06 versus 8.10 without). Moreover, this model is more expensive because $4$ by $4$ patching gives $256$ resolution feature maps.

\begin{figure*}
\centering
\begin{subfigure}[t]{0.33\textwidth}
\centering
\includegraphics[width=.99\textwidth]{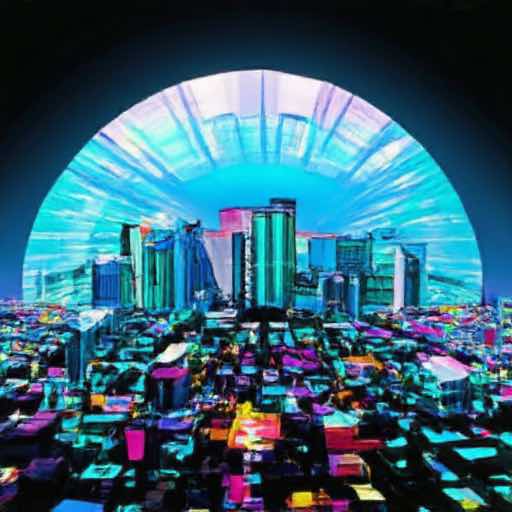}
\caption{\textit{A render of a bright and colorful city under a dome}}
 \end{subfigure} \hfill
\begin{subfigure}[t]{0.33\textwidth}
\centering
\includegraphics[width=.99\textwidth]{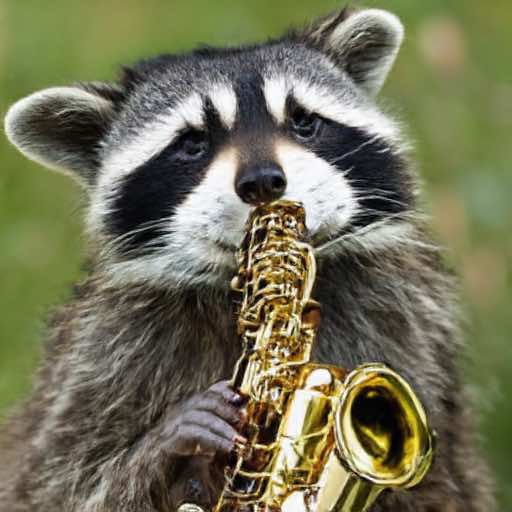}
\caption{\textit{A raccoon playing the saxophone}}
 \end{subfigure} \hfill
\begin{subfigure}[t]{0.33\textwidth}
\centering
\includegraphics[width=.99\textwidth]{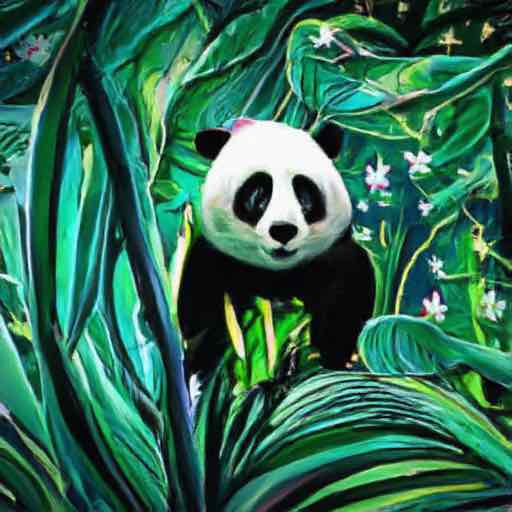}
\caption{\textit{A panda walking through the Jungle, futuristic art}}
\end{subfigure}
\begin{subfigure}[t]{0.33\textwidth}
\centering
\includegraphics[width=.99\textwidth]{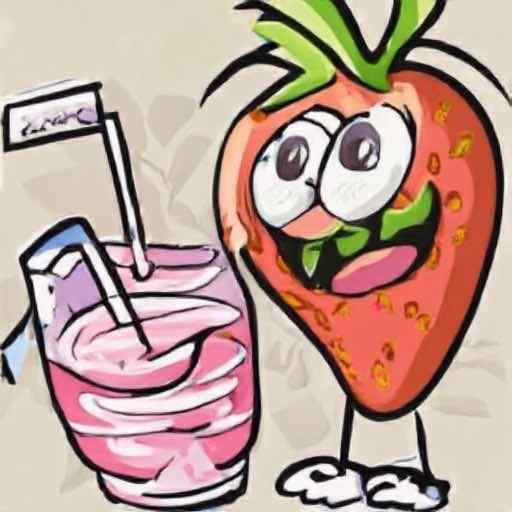}
\caption{\textit{A cartoon of a strawberry drinking a smoothie}}
\end{subfigure} \hfill
\begin{subfigure}[t]{0.33\textwidth}
\centering
\includegraphics[width=.99\textwidth]{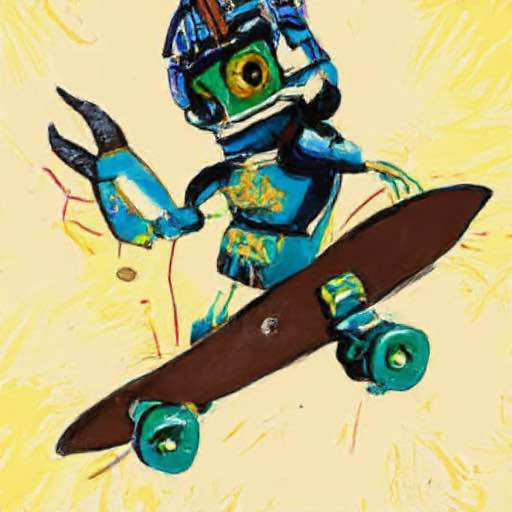}
\caption{\textit{A surrealistic painting of a robot riding a skateboard}}
\end{subfigure} \hfill
\begin{subfigure}[t]{0.33\textwidth}
\centering
\includegraphics[width=.99\textwidth]{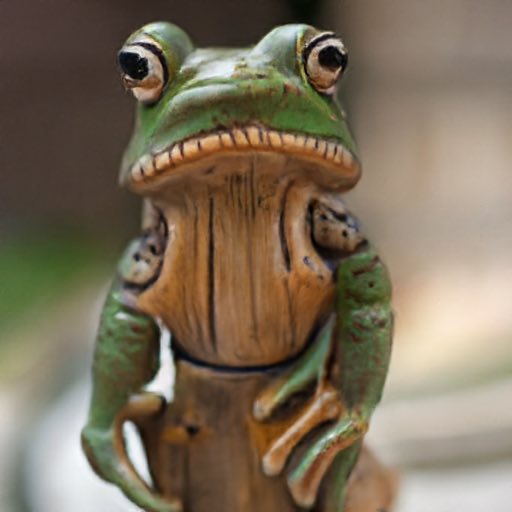}
\caption{\textit{A statue of a frog made of wood}}
\end{subfigure}
\begin{subfigure}[t]{0.33\textwidth}
\centering
\includegraphics[width=.99\textwidth]{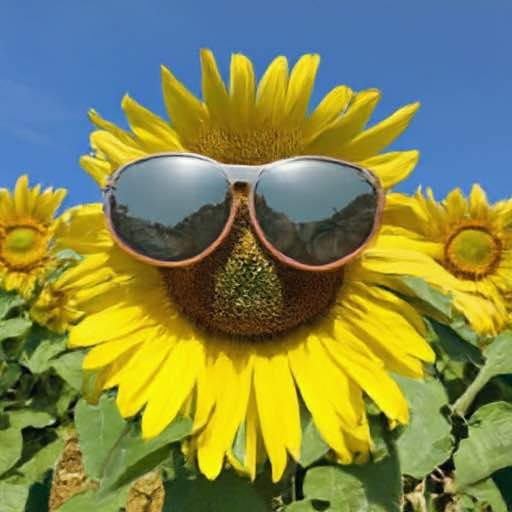}
\caption{\textit{A sunflower wearing sunglasses}}
\end{subfigure} \hfill
\begin{subfigure}[t]{0.33\textwidth}
\centering
\includegraphics[width=.99\textwidth]{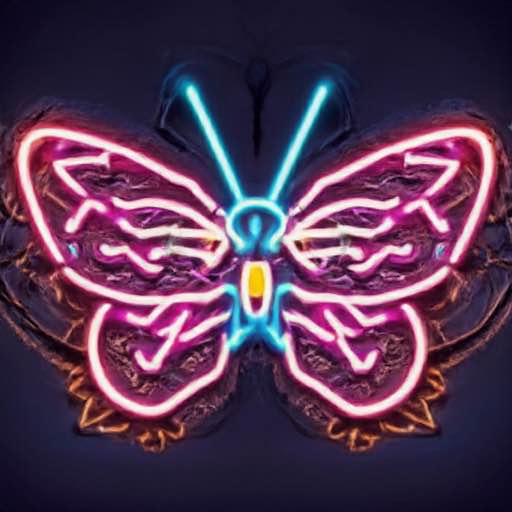}
\caption{\textit{A neon sign of a butterfly}}
\end{subfigure} \hfill
\begin{subfigure}[t]{0.33\textwidth}
\centering
\includegraphics[width=.99\textwidth]{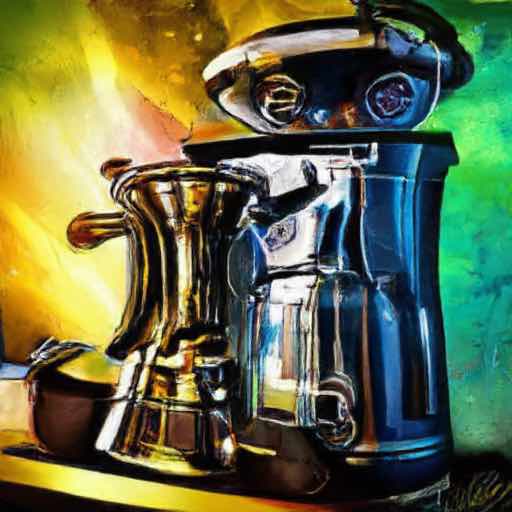}
\caption{\textit{A painting of futuristic coffee machine, vivid colors}}
\end{subfigure}
\caption{Text to image samples at resolution $512 \times 512$. This model was distilled and as a result generating a single image takes 0.42 seconds on a TPUv4 (excluding the text encoder). Similarly, generating a batch of 8 images takes 2.00 seconds.}
\end{figure*}

\begin{figure}
\centering
\begin{subfigure}[b]{0.495\textwidth}
\centering
\includegraphics[width=.23\textwidth]{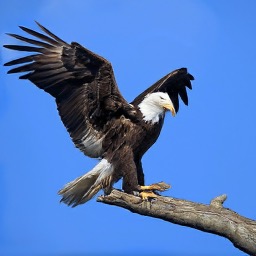} \hspace{.04cm} \includegraphics[width=.23\textwidth]{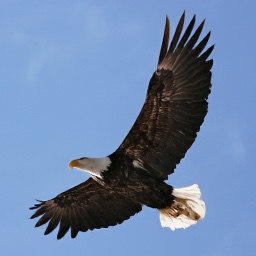} \hspace{.04cm}
    \includegraphics[width=.23\textwidth]{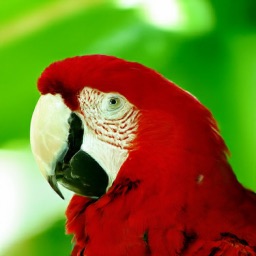} \hspace{.04cm} \includegraphics[width=.23\textwidth]{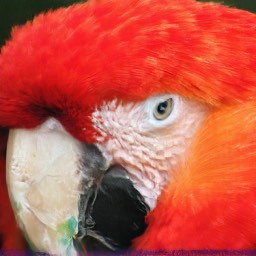} \\ \vspace{.17cm}
\includegraphics[width=.23\textwidth]{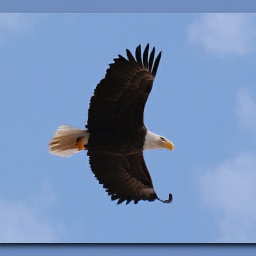} \hspace{.04cm } \includegraphics[width=.23\textwidth]{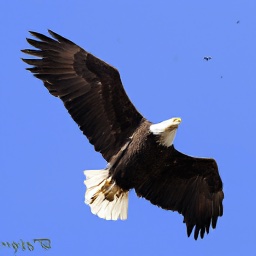} \hspace{.04cm}
    \includegraphics[width=.23\textwidth]{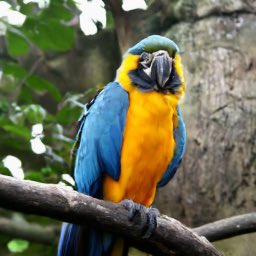} \hspace{.04cm } \includegraphics[width=.23\textwidth]{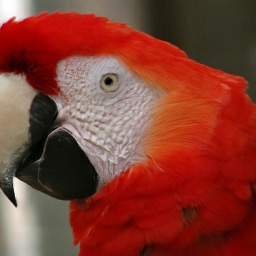} \\ \vspace{.17cm}
\includegraphics[width=.23\textwidth]{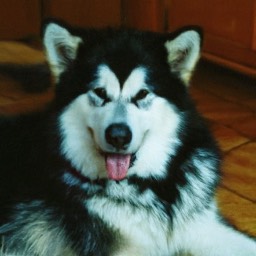} \hspace{.04cm} \includegraphics[width=.23\textwidth]{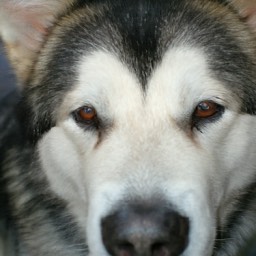} \hspace{.04cm}
    \includegraphics[width=.23\textwidth]{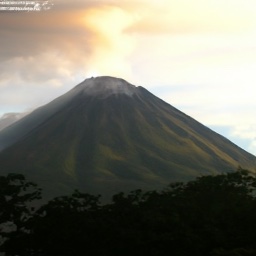} \hspace{.04cm} \includegraphics[width=.23\textwidth]{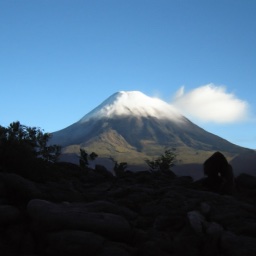} \\ \vspace{.17cm}
\includegraphics[width=.23\textwidth]{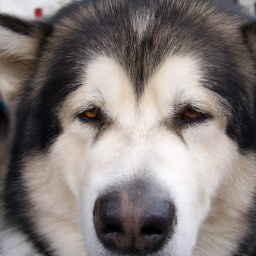} \hspace{.04cm } \includegraphics[width=.23\textwidth]{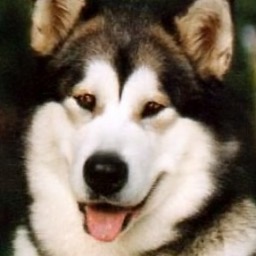} \hspace{.04cm}
    \includegraphics[width=.23\textwidth]{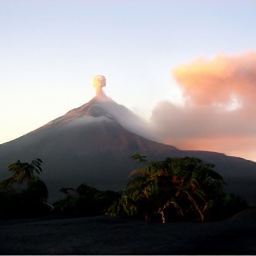} \hspace{.04cm } \includegraphics[width=.23\textwidth]{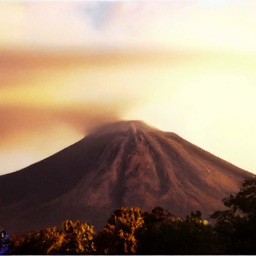} \\ \vspace{.17cm}
\includegraphics[width=.23\textwidth]{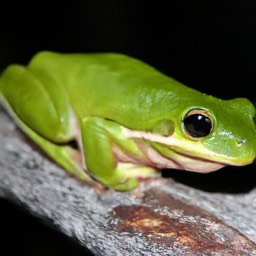} \hspace{.04cm} \includegraphics[width=.23\textwidth]{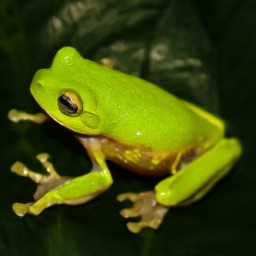} \hspace{.04cm}
    \includegraphics[width=.23\textwidth]{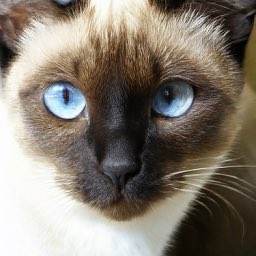} \hspace{.04cm} \includegraphics[width=.23\textwidth]{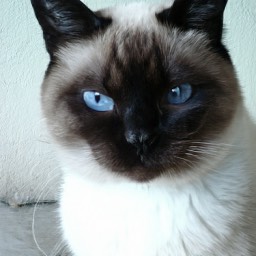} \\ \vspace{.17cm}
\includegraphics[width=.23\textwidth]{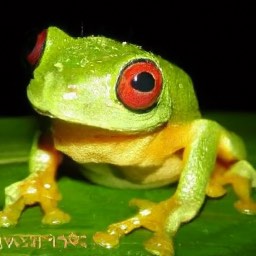} \hspace{.04cm } \includegraphics[width=.23\textwidth]{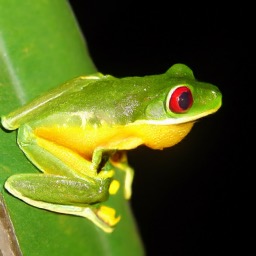} \hspace{.04cm}
    \includegraphics[width=.23\textwidth]{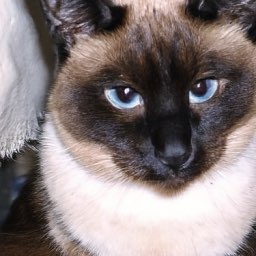} \hspace{.04cm } \includegraphics[width=.23\textwidth]{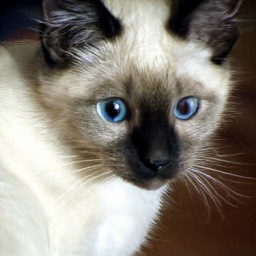} \\ \vspace{.17cm}
\includegraphics[width=.23\textwidth]{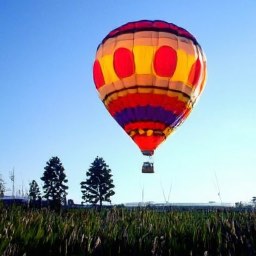} \hspace{.04cm} \includegraphics[width=.23\textwidth]{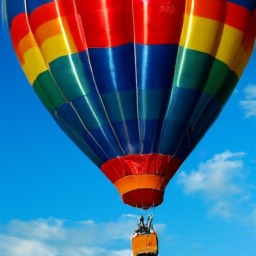} \hspace{.04cm}
    \includegraphics[width=.23\textwidth]{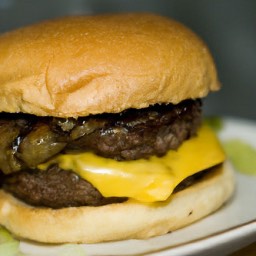} \hspace{.04cm} \includegraphics[width=.23\textwidth]{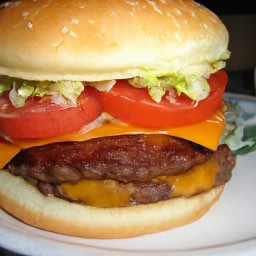} \\ \vspace{.17cm}
\includegraphics[width=.23\textwidth]{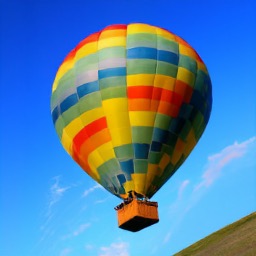} \hspace{.04cm } \includegraphics[width=.23\textwidth]{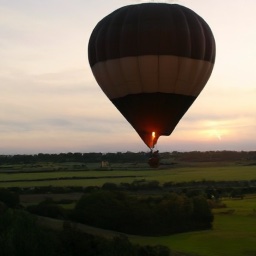} \hspace{.04cm}
    \includegraphics[width=.23\textwidth]{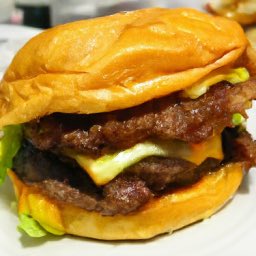} \hspace{.04cm } \includegraphics[width=.23\textwidth]{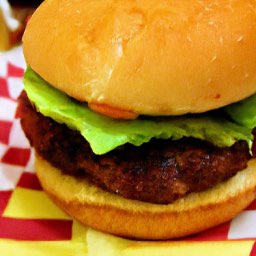} \\ \vspace{.17cm}
\includegraphics[width=.23\textwidth]{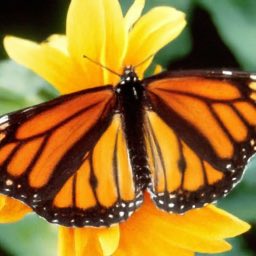} \hspace{.04cm} \includegraphics[width=.23\textwidth]{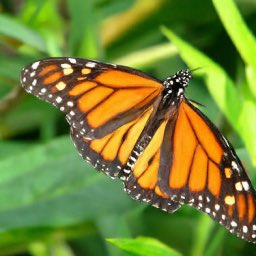} \hspace{.04cm}
    \includegraphics[width=.23\textwidth]{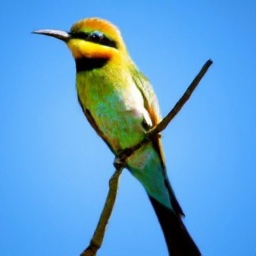} \hspace{.04cm} \includegraphics[width=.23\textwidth]{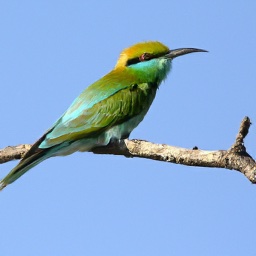} \\ \vspace{.17cm}
\includegraphics[width=.23\textwidth]{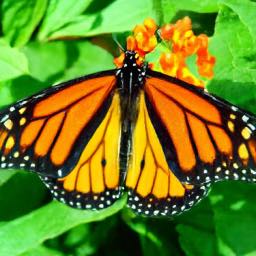} \hspace{.04cm } \includegraphics[width=.23\textwidth]{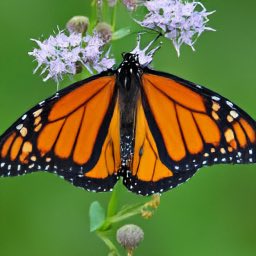} \hspace{.04cm}
    \includegraphics[width=.23\textwidth]{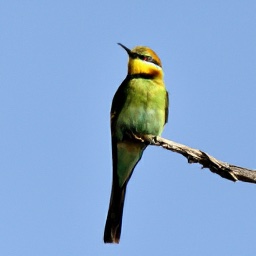} \hspace{.04cm } \includegraphics[width=.23\textwidth]{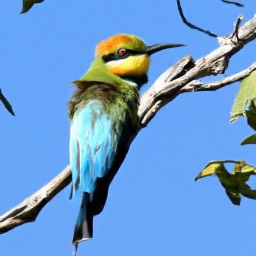} \\ \vspace{.17cm}
\caption{Guidance scale 4}
 \end{subfigure} \hfill
\begin{subfigure}[b]{0.495\textwidth}
\centering
\includegraphics[width=.23\textwidth]{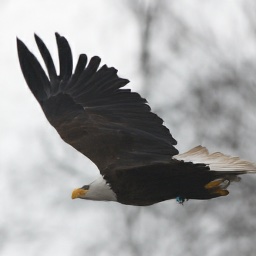} \hspace{.04cm} \includegraphics[width=.23\textwidth]{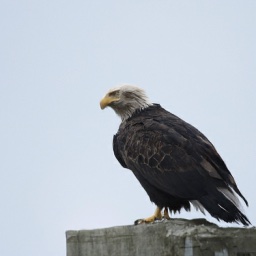} \hspace{.04cm}
    \includegraphics[width=.23\textwidth]{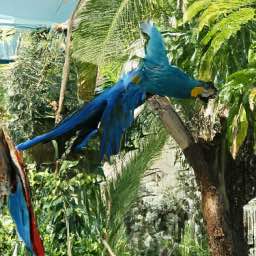} \hspace{.04cm} \includegraphics[width=.23\textwidth]{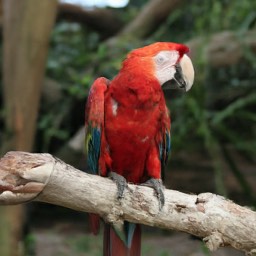} \\ \vspace{.17cm}
\includegraphics[width=.23\textwidth]{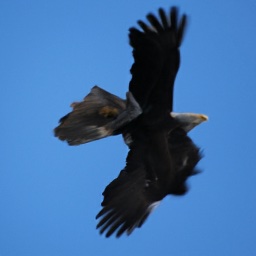} \hspace{.04cm } \includegraphics[width=.23\textwidth]{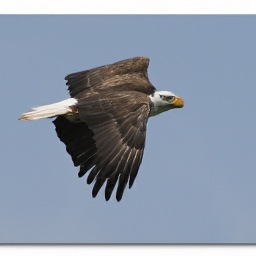} \hspace{.04cm}
    \includegraphics[width=.23\textwidth]{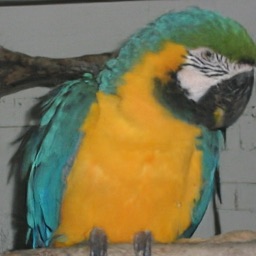} \hspace{.04cm } \includegraphics[width=.23\textwidth]{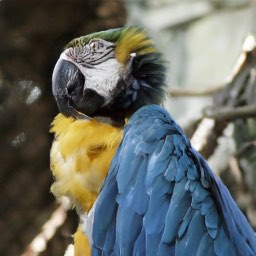} \\ \vspace{.17cm}
\includegraphics[width=.23\textwidth]{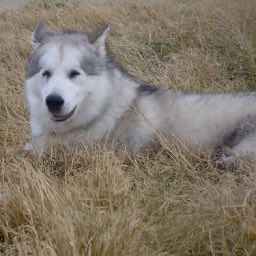} \hspace{.04cm} \includegraphics[width=.23\textwidth]{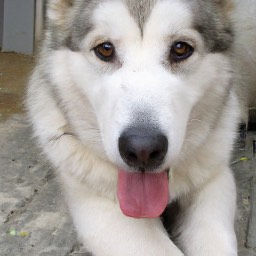} \hspace{.04cm}
    \includegraphics[width=.23\textwidth]{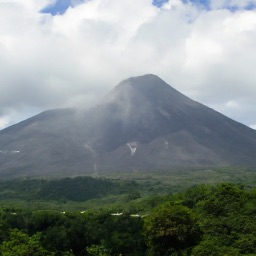} \hspace{.04cm} \includegraphics[width=.23\textwidth]{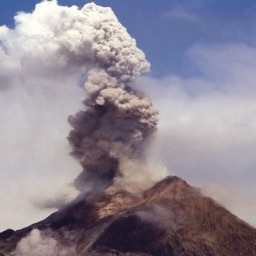} \\ \vspace{.17cm}
\includegraphics[width=.23\textwidth]{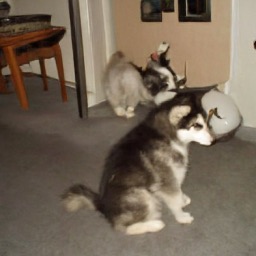} \hspace{.04cm } \includegraphics[width=.23\textwidth]{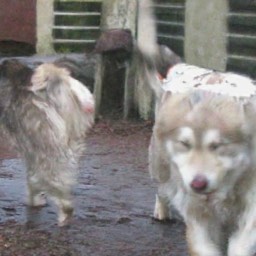} \hspace{.04cm}
    \includegraphics[width=.23\textwidth]{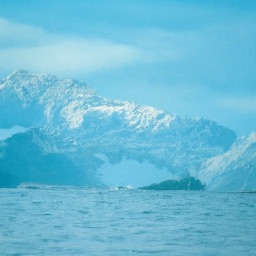} \hspace{.04cm } \includegraphics[width=.23\textwidth]{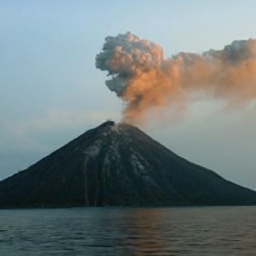} \\ \vspace{.17cm}
\includegraphics[width=.23\textwidth]{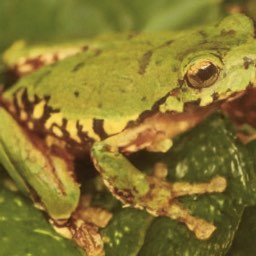} \hspace{.04cm} \includegraphics[width=.23\textwidth]{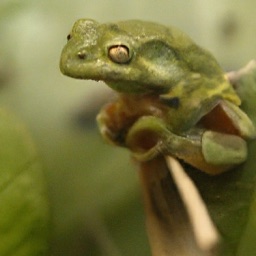} \hspace{.04cm}
    \includegraphics[width=.23\textwidth]{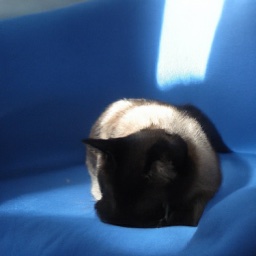} \hspace{.04cm} \includegraphics[width=.23\textwidth]{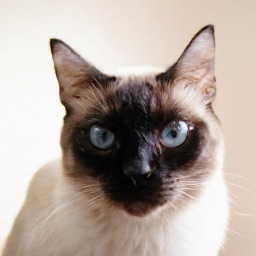} \\ \vspace{.17cm}
\includegraphics[width=.23\textwidth]{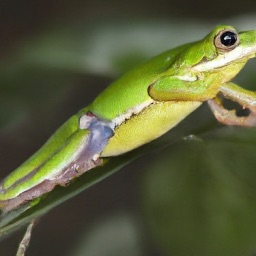} \hspace{.04cm } \includegraphics[width=.23\textwidth]{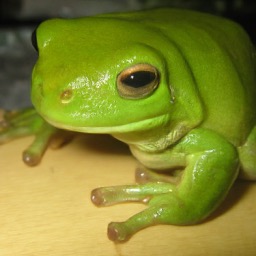} \hspace{.04cm}
    \includegraphics[width=.23\textwidth]{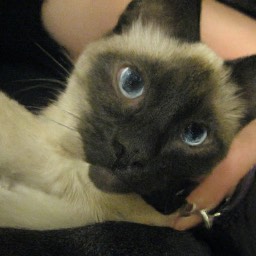} \hspace{.04cm } \includegraphics[width=.23\textwidth]{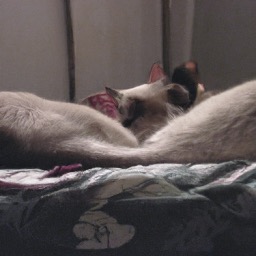} \\ \vspace{.17cm}
\includegraphics[width=.23\textwidth]{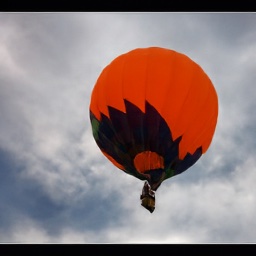} \hspace{.04cm} \includegraphics[width=.23\textwidth]{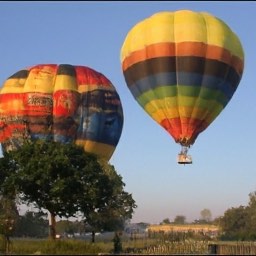} \hspace{.04cm}
    \includegraphics[width=.23\textwidth]{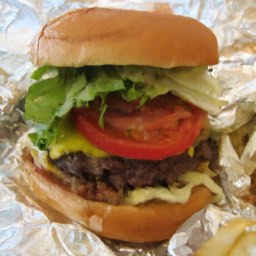} \hspace{.04cm} \includegraphics[width=.23\textwidth]{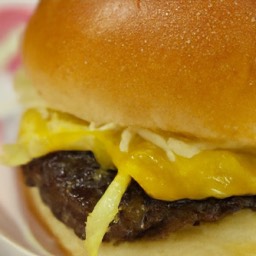} \\ \vspace{.17cm}
\includegraphics[width=.23\textwidth]{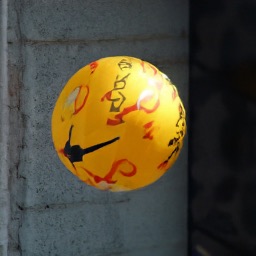} \hspace{.04cm } \includegraphics[width=.23\textwidth]{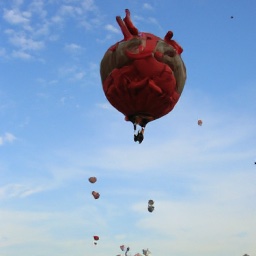} \hspace{.04cm}
    \includegraphics[width=.23\textwidth]{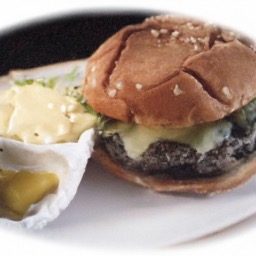} \hspace{.04cm } \includegraphics[width=.23\textwidth]{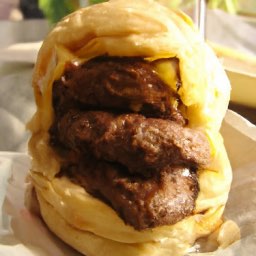} \\ \vspace{.17cm}
\includegraphics[width=.23\textwidth]{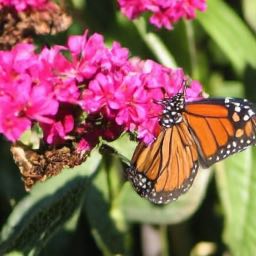} \hspace{.04cm} \includegraphics[width=.23\textwidth]{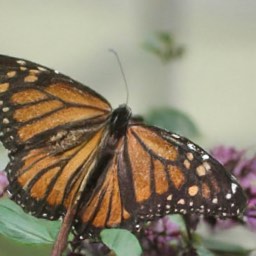} \hspace{.04cm}
    \includegraphics[width=.23\textwidth]{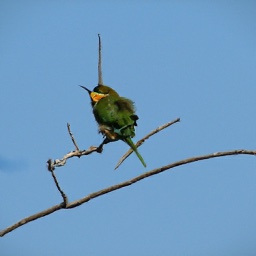} \hspace{.04cm} \includegraphics[width=.23\textwidth]{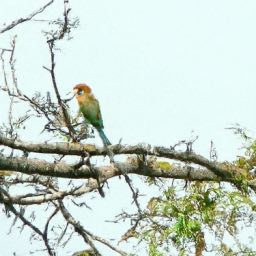} \\ \vspace{.17cm}
\includegraphics[width=.23\textwidth]{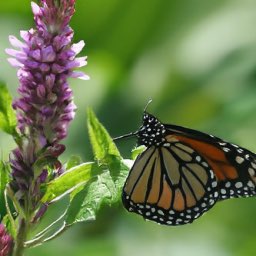} \hspace{.04cm } \includegraphics[width=.23\textwidth]{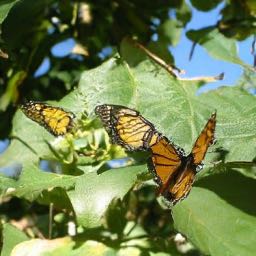} \hspace{.04cm}
    \includegraphics[width=.23\textwidth]{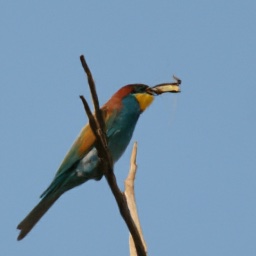} \hspace{.04cm } \includegraphics[width=.23\textwidth]{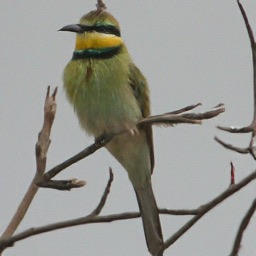} \\ \vspace{.17cm}
\caption{Guidance scale 1 (No guidance)}
 \end{subfigure}
\caption{Random (not cherry picked) samples from the U-ViT on ImageNet 256 $\times$ 256.}
\label{fig:random_samples_imagenet}
\end{figure}

\end{document}